\documentclass[10pt,twocolumn,letterpaper]{article}

\usepackage{iccv}
\usepackage{times}
\usepackage{epsfig}
\usepackage{graphicx}
\usepackage{amsmath}
\usepackage{amssymb}

\usepackage{subcaption}
\usepackage{graphicx}
\usepackage{multirow}
\usepackage{wrapfig}
\usepackage{algorithm}
\usepackage{algpseudocode}

\usepackage{enumitem}
\usepackage[accsupp]{axessibility}  %

\usepackage{amsthm}
\makeatletter
\newtheorem*{rep@theorem}{\rep@title}
\newcommand{\newreptheorem}[2]{%
\newenvironment{rep#1}[1]{%
 \def\rep@title{#2 \ref{##1}}%
 \begin{rep@theorem}}%
 {\end{rep@theorem}}}
\makeatother

\usepackage{amsmath,amsfonts,bm}

\DeclareMathOperator{\Diag}{Diag}

\newcommand{\myparagraph}[1]{\smallskip\noindent\textbf{#1.}}

\def\1{\bm{1}}

\def\vzero{{\bm{0}}}
\def\vone{{\bm{1}}}

\def\vv{{\bm{v}}}
\def\vw{{\bm{w}}}
\def\vx{{\bm{x}}}
\def\vy{{\bm{y}}}
\def\vz{{\bm{z}}}

\def\mC{{\bm{C}}}

\def\mI{{\bm{I}}}

\def\mW{{\bm{W}}}
\def\mX{{\bm{X}}}

\def\mZ{{\bm{Z}}}

\DeclareMathAlphabet{\mathsfit}{\encodingdefault}{\sfdefault}{m}{sl}
\SetMathAlphabet{\mathsfit}{bold}{\encodingdefault}{\sfdefault}{bx}{n}

\def\gM{{\mathcal{M}}}
\def\gN{{\mathcal{N}}}

\def\sR{{\mathbb{R}}}
\def\sS{{\mathbb{S}}}

\def\sZ{{\mathbb{Z}}}

\newcommand\bepsilon{\boldsymbol{\epsilon}}

\newcommand\btheta{\boldsymbol{\theta}}

\newcommand\bGamma{\boldsymbol{\Gamma}}

\newcommand\bPi{\boldsymbol{\Pi}}

\DeclareMathOperator*{\argmin}{arg\,min}

\theoremstyle{plain}

\newreptheorem{theorem}{Theorem}

\theoremstyle{definition}

\newtheorem{problem}{Problem}
\newtheorem{question}{Question}

\theoremstyle{remark}

\usepackage{booktabs}
\usepackage{float}
\usepackage{nccmath}

\usepackage[pagebackref=true,breaklinks=true,letterpaper=true,colorlinks,bookmarks=false]{hyperref}

\hypersetup{
	colorlinks,
	citecolor=blue,
	linkcolor=blue,
	urlcolor=blue}

\usepackage[capitalize]{cleveref}
\crefname{section}{Sec.}{Secs.}
\Crefname{section}{Section}{Sections}
\Crefname{table}{Table}{Tables}
\crefname{table}{Tab.}{Tabs.}
\crefname{problem}{problem}{problems}
\crefname{appendix}{Appendix}{Appendices}

\iccvfinalcopy %

\newcommand{\ours}{MLC}
\newcommand{\mours}{\texttt{\ours}}
\newcommand{\mcr}{\texttt{MCR\textsuperscript{2}}}

\def\iccvPaperID{9756} %

\ificcvfinal\fi

\begin{document}

\title{Unsupervised Manifold Linearizing and Clustering}

\author{\normalsize Tianjiao Ding$^1$ \ \ Shengbang Tong$^2$ \ \ Kwan Ho Ryan Chan$^1$ \ \ Xili Dai$^3$ \ \ Yi Ma$^2$ \ \ Benjamin D. Haeffele$^1$}

\maketitle
\footnotetext[1]{Mathematical Institute for Data Science, Johns Hopkins University, USA $^2$Department of Electrical Engineering and Computer Sciences, University of California, Berkeley, USA $^3$The Hong Kong University of Science and Technology (Guangzhou), PRC}
\setcounter{footnote}{3}

\ificcvfinal\thispagestyle{empty}\fi

\begin{abstract}
We consider the problem of simultaneously clustering and learning a linear representation of data lying close to a union of low-dimensional manifolds, a fundamental task in machine learning and computer vision. When the manifolds are assumed to be linear subspaces, this reduces to the classical problem of subspace clustering, which has been studied extensively over the past two decades. Unfortunately, many real-world datasets such as natural images can not be well approximated by linear subspaces. On the other hand, numerous works have attempted to learn an appropriate transformation of the data, such that data is mapped from a union of general non-linear manifolds to a union of linear subspaces (with points from the same manifold being mapped to the same subspace). However, many existing works have limitations such as assuming knowledge of the membership of samples to clusters, requiring high sampling density, or being shown theoretically to learn trivial representations. In this paper, we propose to optimize the Maximal Coding Rate Reduction metric with respect to both the data representation and a novel doubly stochastic cluster membership, inspired by state-of-the-art subspace clustering results. We give a parameterization of such a representation and membership, allowing efficient mini-batching and one-shot initialization. Experiments on CIFAR-10, -20, -100, and TinyImageNet-200 datasets show that the proposed method is much more accurate and scalable than state-of-the-art deep clustering methods, and further learns a latent linear representation of the data.\footnote{Code is available at {\scriptsize\url{https://github.com/tianjiaoding/mlc}}.}

   \end{abstract}
   
   \begin{figure}[t]

     \centering
     \begin{subfigure}{0.49\linewidth}
       \includegraphics[width=\linewidth,trim={0 1.5cm 0 1.5cm},clip]{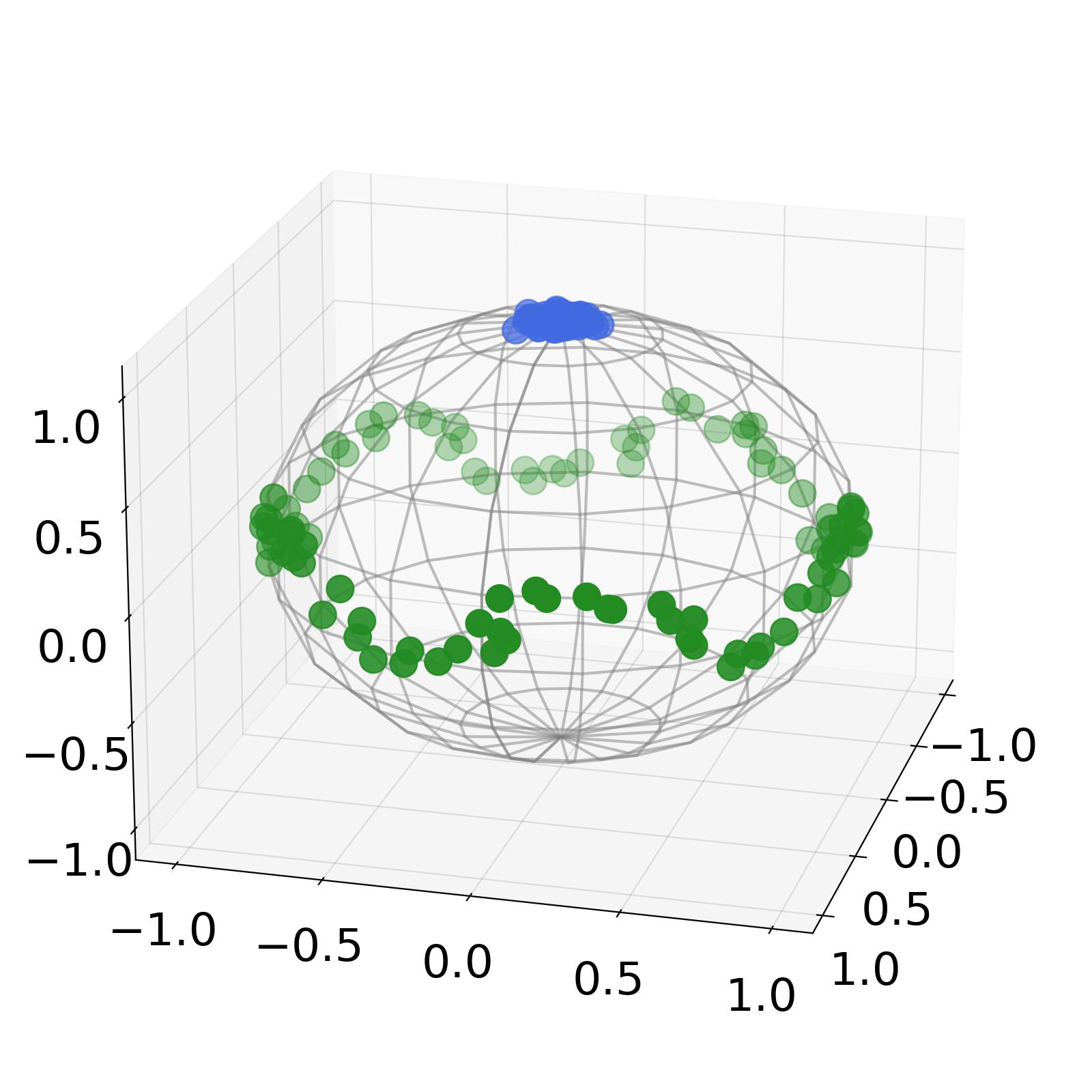}
       \caption{}
       \label{fig:sim1-a}
     \end{subfigure}
     \hfill
     \begin{subfigure}{0.49\linewidth}
       \includegraphics[width=\linewidth,trim={0 1.5cm 0 1.5cm},clip]{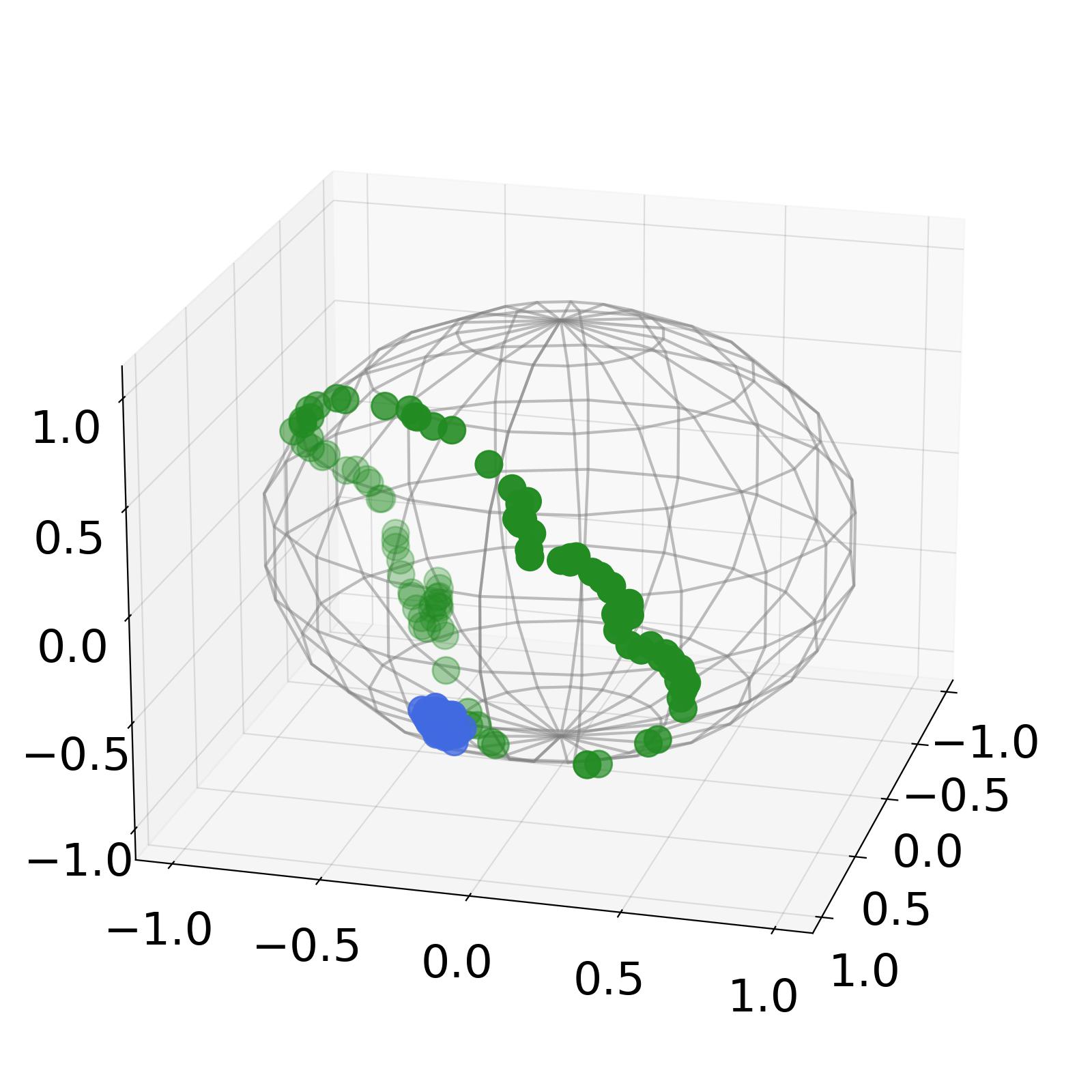}
       \caption{}
       \label{fig:sim1-b}
     \end{subfigure}
     \hfill
     \begin{subfigure}{0.49\linewidth}
       \includegraphics[width=\linewidth,trim={0 1.5cm 0 1.5cm},clip]{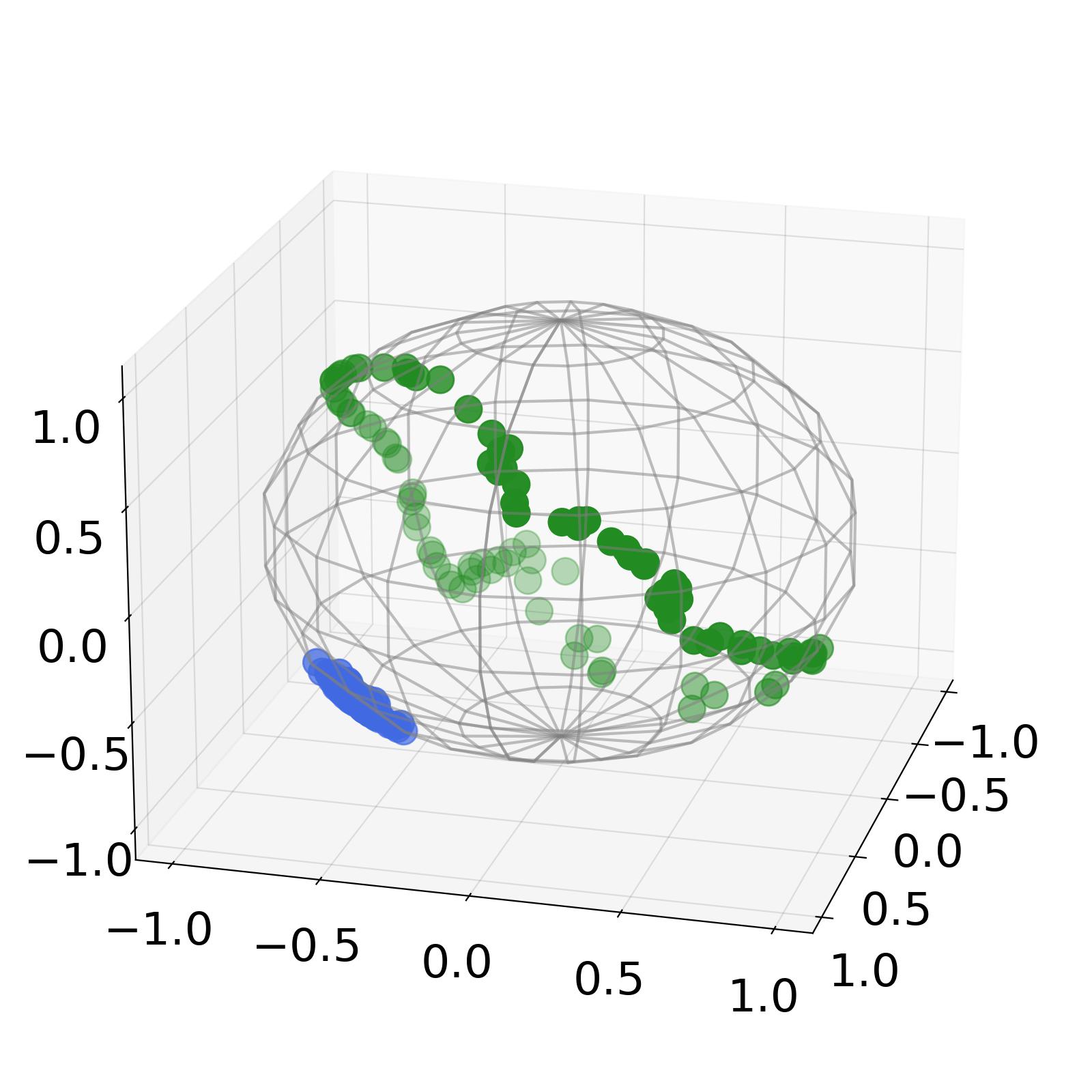}
       \caption{}
       \label{fig:sim1-c}
     \end{subfigure}
     \hfill
     \begin{subfigure}{0.49\linewidth}
       \includegraphics[width=\linewidth,trim={0 1.5cm 0 1.5cm},clip]{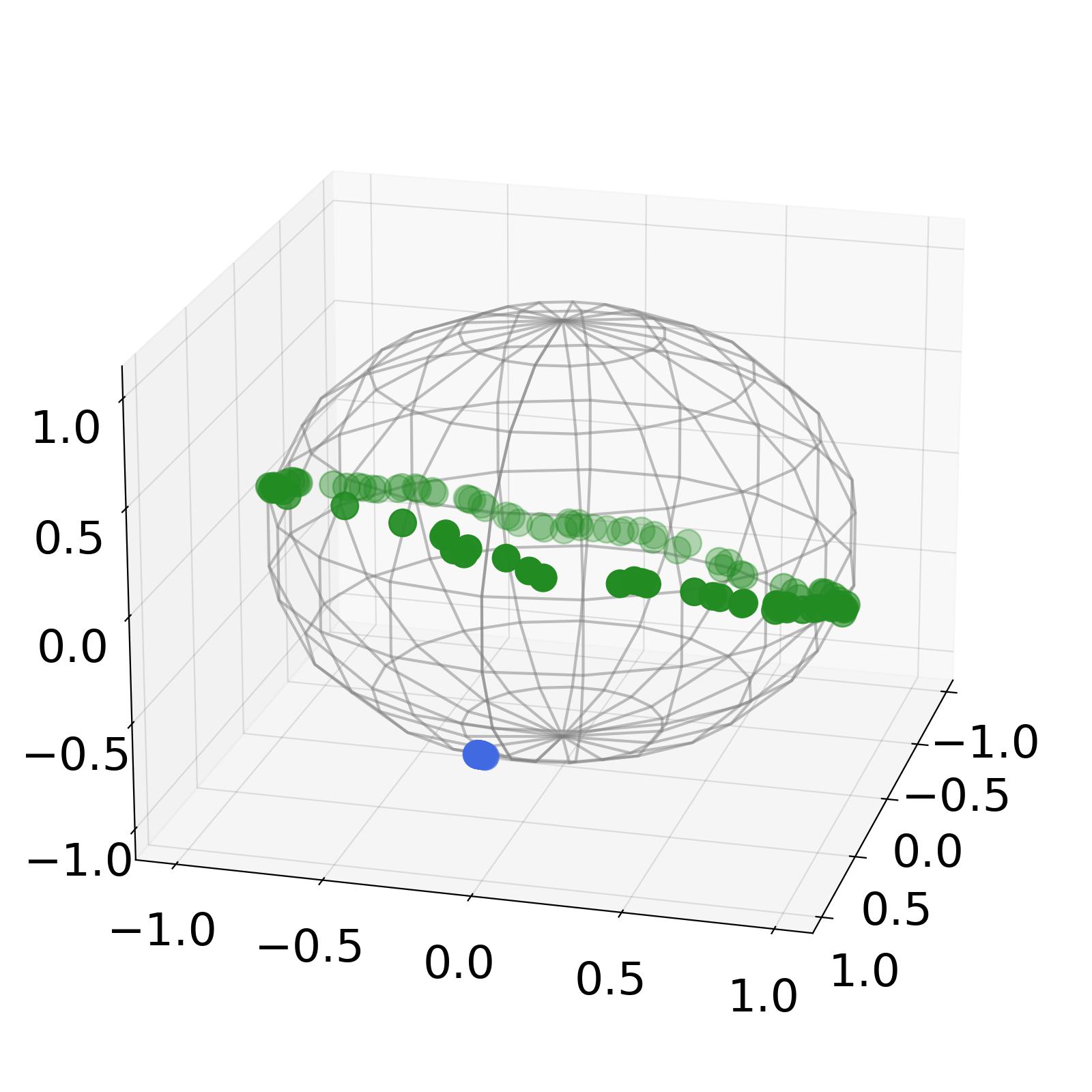}
       \caption{}
       \label{fig:sim1-d}
     \end{subfigure}
     \label{fig:sim1}
     \caption{(a) Input data $\mX$ where $100$ points in green lie on a curve and $100$ in blue lie close to a point. (b) Stage $0$: Features $f_{\boldsymbol{\theta}}(\boldsymbol{X})$ from a neural network $f_{\boldsymbol{\theta}}$ whose parameters $\boldsymbol{\theta}$ are randomly initialized. (c) Stage $1$: Features after self-supervised learning. (d) Stage $2$: Features further improved by the proposed Manifold Linearizing and Clustering (\mours).
     }
   \end{figure}

   \section{Introduction} \label{sec:intro}
   \subsection{Clustering: from Linear to Non-linear Models} \label{sec:intro-sc}
   Clustering is a fundamental problem in machine learning, allowing one to group data into clusters based on assumptions about the geometry of each cluster. As early as the 1950s, the classic $k$-means \cite{Lloyd1957-wk,Forgey1965-xd,Jancey1966-tj,McQueen1967-ru} algorithm emerged to cluster data that concentrate around \textit{distinct centroids}, with numerous
 variants \cite{Bradley1996-cz,Arthur2006-is,Bahmani2012-ip} following. This assumption of distinct centroids was later generalized in \textit{subspace clustering} methods, %
 which aim to cluster data lying close to a union of \textit{low-dimensional linear (or affine) subspaces}\footnote{\label{fn:kmeans-as-subspace}Note, a centroid can be seen as a $0$-dimensional affine subspace.}. This motivated numerous lines of research in the past two decades, leading to various formulations \cite{Elhamifar2009-bw,Elhamifar2013-de,Lu2012-lz,Liu2013-sl,Heckel2015-hm,You2016-ob,Lipor2017-tn,lane2019classifying} with efficient algorithms \cite{You2016-na,You2016-ob,Chen2018-qc} and theoretical guarantees on the correctness of the clustering \cite{Soltanolkotabi2012-kp,Soltanolkotabi2014-rp,Wang2015-yc,Wang2016-gs,Li2018-qr,Tsakiris2018-ej,You2019-kb,Robinson2019-da,Wang2022-rt}. Subspace clustering has been used in a wide range of applications, such as segmenting image pixels \cite{Ma2007-dx,Wang2015-vr,Liu2020-mg}, video frames \cite{Vidal2005-ah,Tierney2014-og,Li2015-wi}, or rigid-body motions \cite[\S 11]{Vidal2016-zp}, along with clustering face images \cite{Georghiades2001-xm,Ho2003-uk,Elhamifar2013-de} or human actions \cite{Wu2016-yv,Gholami2017-nz,Paoletti2021-qh}.

   However, while subspace clustering methods have achieved state-of-the-art performance for certain tasks and datasets, the geometric assumption upon which they rely (namely that the datapoints lie on a union of linear or affine subspaces) is often grossly violated for common high-dimensional datasets. For instance, even in a dataset as simple as MNIST hand-written digits \cite{lecun1998mnist}, images of a single digit do \textit{not} lie close to a low-dimensional subspace; directly applying subspace clustering methods thus fails. Instead, a more natural idea is to assume that each cluster is defined by a \textit{non-linear low-dimensional manifold} and to learn or design a \textit{non-linear transformation} of the data so that points from one manifold are mapped to one linear subspace.  In some cases, one may be able to hand-craft an appropriate transformation of the data, with polynomial or exponential kernel mappings being examples that have been explored in the literature \cite{Elhamifar2011-ca}, and the authors of \cite{Lim2020-sh} show that a subspace clustering method achieves $99\%$ clustering accuracy on MNIST when the data is passed through the scattering-transform \cite{Bruna2013-ru}.

Unfortunately though, hand-crafted design requires one to assume specific and simple families of manifolds %
which is often unrealistic and challenging to apply on complicated data such as natural images. On the other hand, the authors of \cite{Elhamifar2011-ca} propose to cluster data via treating a local neighborhood of the manifold approximately as a linear subspace and applying subspace clustering techniques to local neighborhoods. However, this method requires sufficient sampling density to succeed, which implies a prohibitive number of samples when the manifolds have moderate dimension or curvature. 
    More recently, numerous works propose to learn an appropriate transformation of the data via deep networks and then perform subspace clustering in a latent feature space \cite{Peng2017-gf,Ji2017-yx,Abavisani2018-lt,Zhang2019-yp,Kheirandishfard2020-zc}. Unfortunately, it has been shown that many of these formulations are ill-posed and provably learn trivial representations\footnote{In this paper, we use `representation' and `feature' interchangeably to mean the image of the data under a (learned) transformation.}, with much of the claimed benefit coming from ad-hoc post-processing rather than the method itself \cite{Haeffele2020-kj}. This motivates the one of the primary questions we consider here: 
   
   \begin{question}
       {\em Can we efficiently transform data near a union of low-dimensional manifolds, so that the transformed data lie close to a union of low-dimensional linear subspaces to allow for easy clustering?}
   \end{question}

   \subsection{Learning Diverse and Discriminative Features: from Supervised to Unsupervised Learning}
   Meanwhile, learning a \textit{compact} representation from \textit{multi-modal} data has been a topic of its own interest in machine learning \cite{Bengio2013-ps}. An ideal property of the learned representation is \textit{between-cluster discrimination}, namely, features from different clusters should be well separated, which is often pursued via a loss such as the classic cross-entropy (CE) objective. However, an important yet often ignored property is that the learned representation maintains \textit{within-cluster diversity}. This allows distances of samples within a cluster to be preserved under the learned transformation, which may facilitate downstream tasks such as generation \cite{Dai2022-ui}, denoising \cite{Wang2013-nw}, and semantic interpretation \cite[\S B.3.1]{Yu2020-mx} (see also \S \ref{sec:semantic}). 
    Unfortunately, the representation learned by CE fails to achieve this property and exhibits \textit{neural collapse}, a phenomenon discovered by \cite{Papyan2020-go} with extensive theoretical and empirical analysis \cite{Mixon2020-oe,Zhu2021-ki,Tirer2022-sy,Zhou2022-yq} (even for non-CE objectives \cite{Zhou2022-ob}), where latent features from one cluster tend to collapse to a single point. In contrast, \cite{Yu2020-mx} recently proposed Maximal Coding Rate Reduction (\mcr) as an objective to pursue both of the mentioned ideal properties. In particular, \mcr{} learns a \textit{union-of-orthogonal-subspaces} representation: features from each cluster spread uniformly in a low-dimensional subspace (compact \& within-cluster diverse), and the subspaces corresponding to different clusters are orthogonal to each other (between-cluster discriminative). Nevertheless, \mcr{} requires ground-truth labels to learn such a representation. %
This leads to our second question of interest:
   \begin{question}
       {\em For data lying close to a union of manifolds, can we learn a union-of-orthogonal-subspaces representation, without access to the ground-truth labels?}
   \end{question}

\subsection{Our Contributions}
To address the two interrelated questions, we start with the basic idea of blending the philosophies from \mcr{} and subspace clustering to explore the best of both worlds. This idea leads us to the following contributions.
   \begin{itemize}
       \item \textit{Formulation (\S \ref{sec:formulation})}: We propose Manifold Linearizing and Clustering (\mours) objective \eqref{eq:mcr2-clustering}, which optimizes the \mcr{} loss over both the representation and a novel \textit{doubly stochastic} cluster membership . The latter consists of pair-wise similarities between samples, and it is constrained to be doubly stochastic, inspired by state-of-the-art subspace clustering results \cite{Lim2020-sh,Ding2022-zd}.
       
       \item \textit{Algorithm (\S \ref{sec:algo})}: We describe how to parameterize and initialize the representation and membership, as well as to optimize \mours{} \eqref{eq:mcr2-clustering}. Even though the membership is doubly stochastic, which may appear large in size and hard to constrain, we give an efficient parameterization of it that allows for mini-batching and notably \textit{one-shot initialization}. That is, the membership is initialized with \textit{no additional training} whatsoever leveraging already-initialized representation, which is stable, structured, and efficient. 
        
       \item \textit{Experiments (\S \ref{sec:exp})}: On CIFAR-10, we demonstrate that \mours{} learns a union-of-orthogonal-subspaces representation, and achieves more accurate clustering than state-of-the-art subspace clustering methods. Moreover, on CIFAR-10, -20, -100, and TinyImageNet-200, we show that \mours{} yields \textit{higher clustering accuracy} using \textit{less running time} than state-of-the-art deep clustering methods, even when there are \textit{many or imbalanced clusters}.
\end{itemize}

\subsection{Additional Related Work} \label{sec:related-work}

Beyond the above, we make connections to a few emerging deep-learning-based works related to this paper. 

\myparagraph{Self-supervised Representation Learning}
An important line of research that learns a representation without using ground-truth labels is that of self-supervised learning. It has seen remarkable recent progress thanks to the so-called \textit{joint-embedding} approach. The basic idea of the latter is to learn a representation such that augmentations of the same input have similar features, while features from different inputs do not collapse to a single point. Extending this idea, self-supervised methods such as \texttt{SimCLR} \cite{chen2020simple}, \texttt{BYOL} \cite{grill2020bootstrap}, and \texttt{VICReg} \cite{bardes2021vicreg} are able to learn representations on par with those obtained from supervised learning methods; see \cite{Shwartz-Ziv2023-vh} for an excellent review. Encouraging as it may sound, these methods do not aim to learn a union-of-orthogonal-subspaces representation, nor do they explicitly model clustering in their design.
Nevertheless, we shall see that self-supervised methods are key stepping stones for the proposed method, as they will be used to initialize parts of our model.

\myparagraph{Clustering and Representation Learning} Numerous works have proposed to jointly perform clustering and representation learning, leveraging the success of self-supervised learning. Roughly speaking, most methods consider the following two steps. The first step is to use self-supervised learning to initialize the representation. Indeed, state-of-the-art methods such as \texttt{SCAN} \cite{Van_Gansbeke2020-eo} and \texttt{SPICE} \cite{Niu2022-iq} adopt \texttt{SimCLR} \cite{chen2020simple} and \texttt{MoCoV2} \cite{Chen2020-ws}
as their pre-trained features. Starting from the initial representation, the second step is to iteratively refine the representation and clustering, using the idea of pseudo-labeling \cite{Caron2018-hb,Van_Gansbeke2020-eo,Park2021-jf,Niu2022-iq}.
Despite the promising clustering performance, the representation learned by these methods is not constrained to be both between-cluster discriminative and within-cluster diverse. In contrast, our approach learns a representation with these two ideal properties (Figure \ref{fig:cifar10-ztz}) and also achieves state-of-the-art clustering performance (\Cref{tab:exp-cifar10-20,tab:exp-cifar100-timgnet200,tab:Imbalance_data}). Finally, the work most closely related to this paper is that of Neural Manifold Clustering and Embedding (\texttt{NMCE}) \cite{Li2022-vq} -- we note similarities and differences in terms of formulation, algorithm, and empirical performance at the end of \S \ref{sec:mcr2-unsup-formulation}. 

\begin{table}[]
\centering
\caption{Summary of prior works and our contributions.}
\label{tab:contributions}
\resizebox{\columnwidth}{!}{%
\begin{tabular}{@{}lcccc@{}}
\toprule
                                & \multicolumn{2}{c}{Manifold} & \multicolumn{2}{c}{Self-supervised Initialization} \\ \cmidrule(l){2-5} 
                                & Linearizing   & Clustering   & Representation             & Membership            \\ \midrule
\mcr{} \cite{Yu2020-mx}         & \textbf{yes}  & no           & n/a                        & n/a                   \\
\texttt{SCAN} \cite{Van_Gansbeke2020-eo} & No            & \textbf{yes} & one shot                   & one shot              \\
\texttt{NMCE} \cite{Li2022-vq}           & \textbf{yes}  & \textbf{yes} & one shot                   & no                    \\
\mours{} (Ours)                      & \textbf{yes}  & \textbf{yes} & \multicolumn{2}{c}{\textbf{one shot}}              \\ \bottomrule
\end{tabular}%
}
\end{table}

   \section{Formulation} \label{sec:formulation}
   We begin by making clear the problem of interest. 
   
   \begin{problem}[Unsupervised Manifold Linearizing and Clustering] \label{prb:unsup-manifold-clustering}
   Suppose $\mX = [\vx_1, \ldots, \vx_n] \in \sR^{D \times n}$ is a dataset of $n$ points lying on an union of $k$ low-dimensional manifolds $\bigcup_{j=1}^k\gM_j$. Given $\mX$, we aim to simultaneously 
   \begin{enumerate}
       \item \textit{Cluster the samples}: find $\hat{\vy}$ such that $\vx_i\in \gM_{\hat{y}(i)}$;
       \item \textit{Learn a linear representation}: find a transformation $f: \sR^D \to \sR^d$, such that features $f(\vx_i)$'s from the same cluster spread uniformly in a low-dimensional linear subspace, and the subspaces arising from different clusters are orthogonal.
   \end{enumerate}
  \end{problem}

   In \S \ref{sec:mcr2}, we review the principle of Maximal Coding Rate Reduction (\mcr{}) which is designed to learn ideal representations in a \textit{supervised} manner, i.e., when the ground-truth membership is given. Then in \S \ref{sec:mcr2-unsup-formulation}, we discuss the challenges of simultaneous clustering and learning a representation (i.e., addressing \Cref{prb:unsup-manifold-clustering} in its entirety), for which we propose our \mours{} objective \eqref{eq:mcr2-clustering}. Later in \S \ref{sec:algo}, we further give an algorithm to optimize \mours{} \eqref{eq:mcr2-clustering}.

   \subsection{Supervised Manifold Linearizing via \bf{\mcr{}}} \label{sec:mcr2}
   \newcommand{\Rexp}{R(\mZ;\,\epsilon)}
   \newcommand{\Rcom}[1]{R_c(\mZ,#1;\,\epsilon)}
   \def\txtmcrobj{
   \underbrace{\log\det\left(\mI + \frac{d}{n\epsilon^2} \mZ \mZ^\top \right)}_{\Rexp} - \underbrace{\sum_{j=1}^{k} \frac{\langle \bPi_j, \vone \rangle}{n}\log\det\left(\mI + \frac{d}{\langle \bPi_j, \vone \rangle \epsilon^2} \mZ \Diag(\bPi_j) \mZ^\top \right)}_{\Rcom{\bPi}}
   }
   \def\txtmcrunitnorm{
   \mZ = f_{\btheta}(\mX) 
   }

   When the cluster membership is given as supervision, \mcr{} \cite{Yu2020-mx} is designed to solve part 2) of \Cref{prb:unsup-manifold-clustering}. To begin with, let $f_{\btheta}:\sR^D \to \sS^{d-1}$ be a transformation parameterized by a neural network; this in turn gives  the features $\mZ := [\vz_1, \dots, \vz_n]\in \sR^{d\times n}$ of data with $\vz_i:=f_{\btheta}(\vx_i)\in \sS^{d-1}$. 
   \mcr{} aims to learn an ideal representation by utilizing the coding rate measures. Define the \textit{coding rate}
   \begin{equation}
    \Rexp := \log\det\left(\mI + \frac{d}{n\epsilon^2} \mZ \mZ^\top \right). \nonumber
   \end{equation}
  Intuitively\footnote{ For a rigorous treatment, see for example \cite[\S 2.1]{Ma2007-dx}.}, $\Rexp$ measures some volume of features in $\mZ$ up to $\epsilon>0$ precision, so maximizing it would \textit{diversify} features of all the samples. Likewise, one can apply the measure to features of each cluster and take weighted mean over the clusters; namely, define $\Rcom{\bPi}$ as 
  \begin{align}
    & \sum_{j=1}^{k} \frac{\langle \bPi_j, \vone \rangle}{n}\log\det\left(\mI + \frac{d}{\langle \bPi_j, \vone \rangle \epsilon^2} \mZ \Diag(\bPi_j) \mZ^\top \right) \nonumber.
  \end{align}
  Here $\bPi = [\bPi_1,\dots,\bPi_k]\in \sR^{n\times k}$ is a given \textit{point-cluster} membership such that $\Pi_{ij}=1$ if $\vx_i \in \gM_j$ and $\Pi_{ij}=0$ otherwise, $\vone$ is a vector of all ones so $\langle \bPi_j, \vone \rangle$ is the number of points in cluster $j$, and for any $\vv\in\sR^n$, $\Diag(\vv)$ denotes a diagonal matrix with the entries of $\vv$ along the diagonal. Minimizing $\Rcom{\bPi}$ thus pushes features in each cluster to stay close. \mcr{} learns an ideal representation by \textit{maximizing the coding rate reduction} (hence the acronym)
  \begin{align}
    &\max_{\btheta} \quad  \Rexp-\Rcom{\bPi}  \quad \text{s.t.} \quad  \txtmcrunitnorm  \label{eq:opt-mcr-sup}
   \end{align}
   Notably, it has been shown that given $\bPi$, the features obtained by maximizing the rate reduction has the property that the features of each cluster spread uniformly within a subspace (within-cluster diverse), and subspaces from different clusters are orthogonal (between-cluster discriminative), under relatively mild assumptions \cite[Theorem 2.1]{Yu2020-mx}.

   \subsection{Unsupervised Manifold Linearizing\\ and Clustering} \label{sec:mcr2-unsup-formulation}
   While \mcr{} is designed to learn ideal representations (\S \ref{sec:intro}) when the membership $\bPi$ is given, here we are interested in the \textit{unsupervised} setting where one does \textit{not} have access to membership annotations. To address both parts 1) and 2) of \Cref{prb:unsup-manifold-clustering}, a natural idea is to optimize \mcr{} over both the representation $\mZ$ and membership $\bPi$ via
   \begin{align}
       & \max_{\btheta,\bPi\in\Omega_\circ} \Rexp-\Rcom{\bPi}
       \quad \text{s.t.} \quad\txtmcrunitnorm \label{opt:mcr2-clustering-general}.
   \end{align}
   Here $\Omega_\circ := \{\bPi \in \sR^{n\times k}: \forall i \in [n], \,\, \exists \hat{y}(i) \quad \mathrm{s.t.} \,\,  \Pi_{i\hat{y}(i)}=1 \,\,  \mathrm{and}  \,\, \Pi_{ij}=0 \,\, \mathrm{for} \,\, j\neq  \hat{y}(i)\}$ is the set of all `hard' assignments, i.e., each row of $\bPi$ is a one-hot vector. However, this optimization is in general combinatorial: its complexity grows exponentially in $n$ and $k$, and it does not allow smooth and gradual changes of $\bPi$. Further, a second challenge is the chicken-and-egg nature of this problem: If one already has an ideal representation $\mZ$, then existing subspace clustering methods can be applied on $\mZ$ to estimate the membership. Likewise, if one is given the ground-truth membership $\bPi$ of clusters, then solving (\ref{eq:opt-mcr-sup}) would lead to an ideal representation. However, the $\mZ$ and $\bPi$ at the beginning of optimization are typically far from ideal. 
   
   In the rest of this section, we propose a so-called \textit{doubly stochastic membership} to deal with the combinatorial challenge. To tackle the chicken-and-egg nature, we parameterize and initialize the representation and membership in an efficient and effective way, as we will discuss in \S \ref{sec:algo}.
   
   \myparagraph{Doubly Stochastic Subspace Clustering} To address the combinatorial nature of estimating the memberships, we draw inspiration from the closely related problem of \textit{subspace clustering}, where the goal is to cluster $n$ samples assumed to lie close to a union of $k$ low-dimensional subspaces (\S \ref{sec:intro-sc}). In this case, one typically does not directly learn an $n\times k$ matrix denoting memberships of $n$ points into $k$ subspaces. Instead, one first learns an affinity matrix $\bGamma\in\sR^{n\times n}$ signaling the similarities between pairs of points, and then applies spectral clustering on the learned $\bGamma$ to obtain a final clustering \cite{Elhamifar2009-bw,Elhamifar2013-de,Lu2012-lz,Liu2013-sl,Heckel2015-hm,You2016-ob}. In particular, requiring doubly-stochastic constraints on the affinity $\bGamma$ is shown theoretically to suppress false inter-cluster connections for clustering \cite{Ding2022-zd} along with state-of-the-art empirical performance for subspace clustering \cite{Lim2020-sh}.
   
   Inspired by the above, we propose a constraint set $\Omega$ for matrix $\bGamma$ to be the set of $n\times n$ doubly stochastic matrices, 
   \begin{equation}
       \Omega = \{\bGamma \in \sR^{n\times n}: \bGamma \geq 0, \quad \bGamma \vone = \bGamma^\top \vone = \vone\}.
   \end{equation}
   However, this constraint alone is insufficient for strong clustering performance: Consider optimizing \mcr{} with respect to $\bGamma\in\Omega$ only, and note that the objective is convex with respect to $\bGamma$. Since we maximize a convex function with respect to convex constraints $\Omega$, an optimal $\bGamma$ would lie at an extreme point of $\Omega$, which for doubly stochastic matrices is a permutation matrix. This is not ideal for clustering, as it implies that every point is assigned to its own distinct cluster, and there is no incentive to merge points into larger clusters.  
   To resolve this issue, we use a negative entropy regularization $\sum_{ij} \Gamma_{ij} \log(\Gamma_{ij})$ to $\bGamma$ which biases $\bGamma$ toward the uniform matrix $\tfrac{1}{n} \1 \1^\top$. By tuning the coefficient of such regularization, we can also tune the sparsity level of $\bGamma$. This regularization can be conveniently integrated into the network architecture, as we will see in \S \ref{sec:algo}.  Now we are ready to state our proposed formulation Manifold Linearizing and Clustering (\mours{}): 
   \def\txtmcrseobj{
   \underbrace{\log\det\left(\mI + \frac{d}{n\epsilon^2} \mZ \mZ^\top \right)}_{\Rexp} - \underbrace{\frac{1}{n} \sum_{j=1}^{n} \log\det\left(\mI + \frac{d}{\epsilon^2} \mZ \Diag((\bGamma)_j) \mZ^\top \right)}_{\Rcom{\bGamma}} 
   }
   \def\txtmcrds{
   \bGamma = h_{\btheta}(\mX) \in \Omega
   }

   \begin{align}
    & \max_{\btheta} \quad \Rexp-\Rcom{\bGamma}  \label{eq:mcr2-clustering} \\
    & \quad \text{s.t.} \quad  \txtmcrunitnorm, \,\, \txtmcrds, \text{ where } \nonumber\\
      & \medmath{ \Rcom{\bGamma} = \frac{1}{n} \sum_{j=1}^{n} \log\det\left(\mI + \frac{d}{\epsilon^2} \mZ \Diag((\bGamma)_j) \mZ^\top \right)} \nonumber.
   \end{align}
   Note that both $\mZ$ and $\bGamma$ are parameterized by neural networks. While a doubly stochastic membership $\bGamma$ may seem large in size and hard to constrain, we explain in \S \ref{sec:algo} how we parameterize $\bGamma$ so that the constraints are satisfied by construction and efficient mini-batching is allowed.

   \begin{figure*}[t]
   \centering\includegraphics[width=0.9\linewidth,trim={0 3cm 0 1cm},clip]{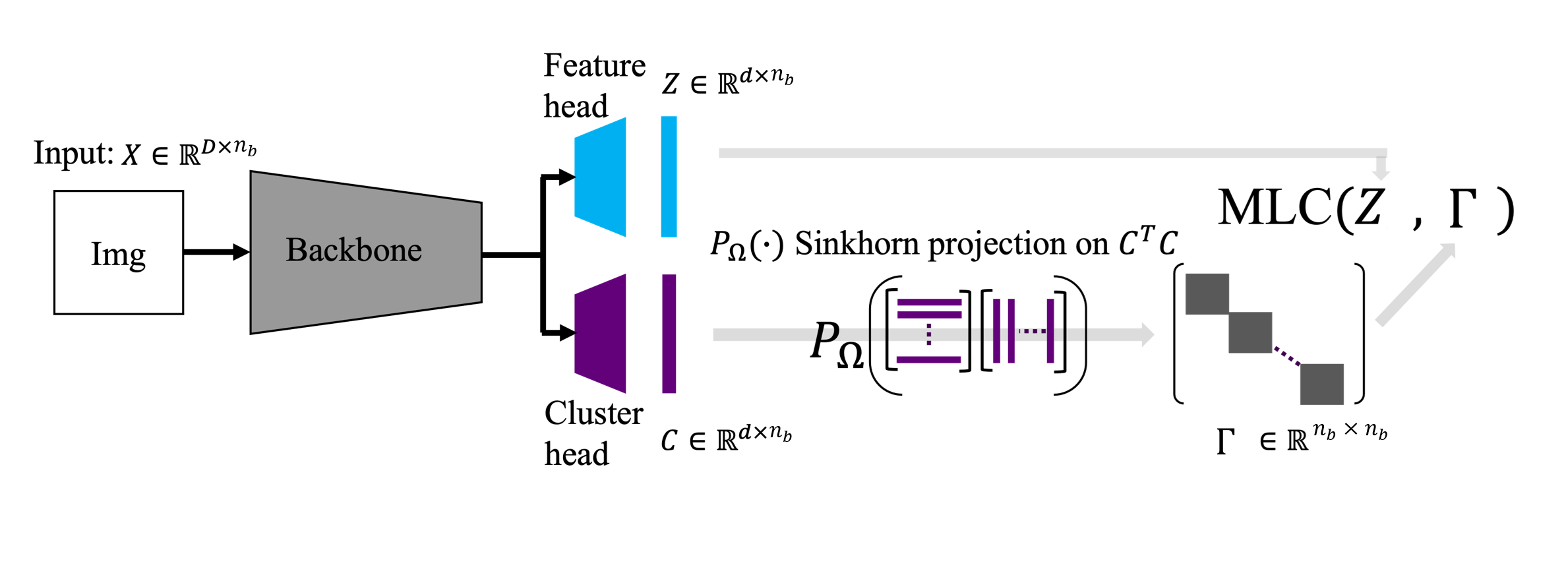}
   \caption{Overall architecture for optimizing the proposed Manifold Linearizing and Clustering (\mours{}) objective \eqref{eq:mcr2-clustering}. Given a mini-batch of $n_b$ input samples $\mX$ each lying in $\sR^D$, their $d$-dimensional representation is given by $\mZ$
   . Further, their doubly stochastic membership matrix $\bGamma$ is given by taking an inner product kernel of the output of the cluster head $\mC$ followed by a doubly stochastic projection.}
   \label{fig:arch}
   \end{figure*}

   \myparagraph{Comparison with \texttt{NMCE}} A recent paper (\texttt{NMCE}) \cite{Li2022-vq} studies the same \Cref{prb:unsup-manifold-clustering} as in this paper, and also proposes to optimize \mcr{} over both the representation and membership. Their method adopts an $n\times k$ matrix $\bPi$ to model the \textit{point-cluster} membership; in contrast \mours{} uses a \textit{doubly stochastic point-point} membership $\bGamma$ inspired from the state-of-the-art subspace clustering (as stated above). Although seemingly not particularly significant, we note that in practice this allows for significantly simpler initialization strategies since we can initialize with a $n \times n$ estimate of an affinity matrix rather than a $n \times k$ estimate of cluster membership.  We further elaborate on algorithmic differences at the end of \S \ref{sec:algo-init}, and give empirical evidence that \mours{} is more accurate (\Cref{tab:exp-cifar10-20}) and stable against randomness (\S \ref{sec:appendix-nmce}).

   \section{Algorithm} \label{sec:algo}
   In this section, we describe how to parameterize the representation $\mZ$ and doubly stochastic membership $\bGamma$ (\S \ref{sec:algo-param}), as well as how to initialize them (\S \ref{sec:algo-init}) -- in an efficient and effective manner. We summarize the meta-algorithm in \Cref{alg:mcr2-clustering}, and the overall architecture in \Cref{fig:arch}.

   \subsection{Efficient Parameterization} \label{sec:algo-param}
   \myparagraph{Parameterizing $\mZ$} As is common practice, we take an existing network architecture such as ResNet-18 as the backbone. We append a few affine layers with non-linearities as the \textit{feature head} to further transform the output of the backbone to $\sR^{d}$, followed by a projection layer to respect the unit sphere $\sS^{d-1}$ constraint.

   \myparagraph{Parameterizing $\bGamma$} Different from parameterizing $\mZ$, this is much less trivial: If one were to directly take $\bGamma$ as decision variables in $\Omega$, it would lead to maintaining $O(n^2)$ variables, which is prohibitive for large datasets (e.g., $n=10^6$ for ImageNet). To allow efficient computation, we again draw inspiration from \textit{subspace clustering}: There, the membership $\bGamma$ given data $\mX$ often takes the form of $g(\mX)^\top g(\mX)$ for some transformation $g$, such as in the inner product kernel \cite{Heckel2015-hm,Ding2022-zd} where $g(\mX)=\mX$ or the least square regression \cite{Lu2012-lz} where $g(\mX)=(\mI+\lambda\mX^\top\mX)^{-1/2}\mX$. It is then tempting to take a neural network $g_{\btheta}$ and use  $\mC^\top \mC$ as the membership where $\mC=g_{\btheta}(\mX)$. Nevertheless, such a matrix is in general \textit{not} doubly stochastic, i.e., $\mC^\top \mC\notin\Omega$. To obtain a doubly stochastic membership, we further apply a Sinkhorn projection layer $P_{\Omega, \eta}(\cdot)$ \cite{Sander2021-tl,Eisenberger2022-oo}, which gives $\bGamma=P_{\Omega, \eta}(\mC^\top \mC)\in \Omega$, where $\eta$ is the coefficient of entropy regularization.\footnote{So in \eqref{eq:mcr2-clustering}, $\bGamma=h_{\btheta}(\mX)$ is simply $\bGamma=P_{\Omega, \eta}(g_{\btheta}(\mX)^\top g_{\btheta}(\mX)).$ 
   } As in parameterizing $\mZ$, we implement $g_{\btheta}$ by taking the same backbone and appending layers of the same type to be the \textit{cluster head}. As we shall see soon in \S \ref{sec:algo-init}, such a parameterization further allows us to initialize both $\mZ$ and $\bGamma$ in \textit{one shot} using self-supervised learning.

   \myparagraph{Complexity} Thanks to the above parameterization, we can do forward and backward passes efficiently via mini-batches. For a mini-batch of $n_b$ samples ($n_b \ll n$ typically), the mini-batched versions of $\mZ$, $\mC$ and $\bGamma$ have sizes $d\times n_b$, $d\times n_b$ and $n_b\times n_b$, respectively (\Cref{fig:arch}).\footnote{Interested readers are referred to \cite{Baek2022-bj} for efficient computation of $\log\det$ via variational forms, and \cite{Eisenberger2022-oo} for efficient Sinkhorn iterations via implicit differentiation; these are beyond the scope of this paper.}

   \subsection{Efficient Initialization} \label{sec:algo-init}
   Since the proposed \mours{} objective \eqref{eq:mcr2-clustering} is non-convex, it is important to properly initialize both $\mZ$ and $\bGamma$ to converge to good (local) minimum. 

   \myparagraph{Initializing $\mZ$: Self-supervised Representation Learning} Randomly initialized features could be far from being ideal (in the sense defined in \S \ref{sec:intro}), and further may not respect the invariance to augmentation, i.e., the augmented samples should have their representation close to each other. Thus, we initialize the features using a self-supervised learning called \textit{total coding rate} (\texttt{TCR}) \cite{Li2022-vq}
   \def\txttcrobj{
     R\Big(\frac{\mZ+\mZ'}{2};\,\epsilon\Big) + \lambda \sum_{i=1}^{n} |\vz_i^\top \vz'_i|
   }
   \def\txttcrunitnorm{
   \vz_i',\vz_i\in \sS^{d-1}, \quad \forall i\in [n]
   }
   \begin{align}
       & \max_{\btheta} \quad  \txttcrobj, \nonumber\\
       & \quad \text{s.t.} \quad  \txttcrunitnorm \label{eq:opt-tcr-sup},
   \end{align}
   where for every $i$, $\vz_i$ and $\vz'_i$ are features of different augmentations of the $i$-th sample. 
   This essentially requires that features from different augmentations of the same sample should be as close as possible, whereas features from different samples should be as uncorrelated as possible. \footnote{\texttt{TCR} is introduced here since it also uses coding rate measures. In principle, many contrastive learning methods could be used for initializing the features, e.g., as we demonstrate via experiments (\Cref{tab:exp-cifar10-20}).}

   \myparagraph{Initializing $\bGamma$} An ideal initialization of $\bGamma$ would be such that if $\bGamma_{ij}$ has a high value then points $i, j$ are likely to be from the same true cluster and vice versa. Luckily, after the self-supervised feature initialization mentioned above, $\mZ$ already have some structures which we can utilize. Thus, we propose to initialize $\bGamma$ with $P_{\Omega, \eta}(\mZ^\top \mZ)$; this is easily implemented by copying the parameters from the feature head to the cluster head \textit{once} after the self-supervised initialization of the features, thanks to the parameterization of $\bGamma$ discussed in \S \ref{sec:algo-param}. 

   With the parameterization and initialization of our doubly stochastic membership $\bGamma$ set up, we are ready to contrast it with a popular alternative in the sequel.

   \myparagraph{Doubly Stochastic Membership vs. Point-Cluster Membership} 
   Different from the doubly stochastic \textit{point-point} membership $\bGamma$ proposed in this paper, prior deep representation learning and clustering works \cite{Van_Gansbeke2020-eo,Li2022-vq,Niu2022-iq} often model a \textit{point-cluster} membership. That is, an $n\times k$ matrix $\bPi$ where each row represents the probability of a point belonging to $k$ clusters. $\bPi$ is parameterized by a neural network (or a cluster head if one wishes), initialized randomly or otherwise via an extra training stage after the representation $\mZ$ is initialized. We highlight a few advantages of using a doubly stochastic point-point membership over a point-cluster one: 
   \begin{itemize}[wide,parsep=0pt,topsep=0pt]
    \item \textit{Stable}: As $\bGamma$ is initialized deterministically, the performance of \mours{} is more stable compared to incorporating randomness in initializing the cluster head. We further justify this point empirically in \S \ref{sec:appendix-nmce}.
    \item \textit{Structured}: Initialization of $\bGamma$ takes advantage of structures from self-supervised initialized $\mZ$. As a side benefit, \mours{} automatically gains from developments of self-supervised representation learning.
    \item \textit{Efficient}: Once $\mZ$ is initialized, $\bGamma$ can be initialized with no additional cost whatsoever, compared to using any extra training stage to initialize the cluster head. In contrast, e.g., \texttt{SCAN} \cite{Van_Gansbeke2020-eo} trains the cluster head with $10$ different random initializations, which is time-consuming.
   \end{itemize}
   
   \myparagraph{Data Augmentation} Beyond initializing $\mZ$, it is often desirable to incorporate augmentation in optimizing the \mours{} objective \eqref{eq:mcr2-clustering}. Specifically, from $\{\mX^{(a)}\in \sR^{D\times n} \}_{a=1}^A$ the dataset $\mX$ under $A$ different augmentations, one computes $(\mZ^{(a)}\in\sR^{d\times n}, \bGamma^{(a)}\in\sR^{d\times n})$ for each augmentation $a$, and use in \eqref{eq:mcr2-clustering}
   {\small
   \begin{equation}
       \mZ=P_{\sS^{d-1}}\left(\frac{1}{A} \sum_{a=1}^A \mZ^{(a)} \right), \quad \bGamma=\frac{1}{A}\sum_{a=1}^A\bGamma^{(a)}\in \Omega.\label{eq:aug}
   \end{equation}
   }
   Note that one can benefit from parallelization by putting $\mX^{(a)},\mZ^{(a)},\bGamma^{(a)}$ for each augmentation $a$ on one computing device, since $\bGamma^{(a)}$ only depends on $\mX^{(a)}$ but not from other augmentations.
   \begin{algorithm}[t]
   \caption{\mours: Manifold Linearizing and Clustering}\label{alg:mcr2-clustering}
   \begin{algorithmic}[1]
   \Require $\mX\in \sR^{D\times n}$, \,\, $\epsilon,\eta > 0$, \,\, $d,k,n_b,T,A \in \sZ_{\geq 0}$
   \State initialize $\mZ$ by self-supervised representation learning \Comment{(\ref{eq:opt-tcr-sup})} \label{alg-line:self-sup-init}
   \State initialize $\bGamma$ via parameter copying
       \For{$t=1,\dots,T$} \label{alg-line:start-mlc}
           \State  $\bar{\mX}\in\sR^{D\times n_{b}} \gets$ sample a batch from $\mX$
           \State $\bar{\mX}^{(1)},\dots, \bar{\mX}^{(A)}\gets$ apply $A$ augmentations to $\bar{\mX}$
           \State $\bar{\mZ},\bar{\bGamma}$ $\gets$ forward pass with $\{\bar{\mX}^{(a)}\}_{a=1}^A$ and network parameters $\btheta$ \Comment{(\ref{eq:aug})}
           \State $\nabla_{\btheta}$(\ref{eq:mcr2-clustering}) $\gets$ backward pass
           \State $\btheta\gets$update $\btheta$ using some optimizer on $\nabla_{\btheta}$(\ref{eq:mcr2-clustering})
       \EndFor \label{alg-line:end-mlc}
   
   \State run spectral clustering on $\bGamma$ to estimate labels $\hat{\vy}$ \label{alg-line: spectral-clustering}
   \Ensure $\mZ, \hat{\vy}$
   \end{algorithmic}
   \end{algorithm}

   \section{Experiments on Real-World Datasets} \label{sec:exp}
    We empirically verify that \mours{} learns a union-of-orthogonal-subspaces representation, and yields more accurate clustering than state-of-the-art subspace clustering methods (\S \ref{sec:exp-sc}). 
   Further, we show that \mours{} outperforms state-of-the-art deep clustering methods, even when there are many or imbalanced clusters (\S \ref{sec:exp-deep}).
   
   \myparagraph{Metrics} To evaluate the clustering quality, we run spectral clustering on learned membership matrix $\bGamma$, and report the normalized mutual information (NMI, \cite{Strehl2002-wx}) and clustering accuracy (ACC, \cite{Lee2015-lv}), as are commonly used in clustering tasks. To evaluate the learned representation, we define the following metric: for a collection of points  $\mW = [\vw_1, \ldots, \vw_l]\in \sR^{d \times l}\,\, (l>d)$ with associated singular values $\{\sigma_i\}_{i=1}^d$,
   define the numerical rank of $\mW$ as $\argmin_r\left\{r: \sum_{i=1}^r\sigma_i^2 / \sum_{i=1}^d\sigma_i^2>0.95\right\}$. Now, one can measure the numerical rank of the learned representation $\mZ$, as well as that of each ground-truth cluster\footnote{They are defined by the true membership, so that the numerical rank metric is decoupled from the quality of learned membership $\bGamma$.} of $\mZ$. A low numerical rank of $\mW$ implies that points in $\mW$ lie close to a low-dimensional subspace. We further report the cosine similarity of learned representation, which is simply $|\vz_i^\top \vz_j|$ for points $i$ and $j$, since $\|\vz_i\|=1$ by construction in \eqref{eq:mcr2-clustering}. Finally, to compare the efficiency of methods we report the training time in \S \ref{sec:exp-deep}, where the experiments are run on $2$ Nvidia RTX3090 GPUs.
   
   \begin{figure*}[t!]
    \begin{minipage}[b]{0.66\textwidth}
      \centering
      \small
      \subfloat[Coding rate of all features $R$, that of clustered features $R_c$, and the rate reduction $\Delta R=R-R_c$.\label{fig:cifar10-a}]{\includegraphics[width=0.49\linewidth]{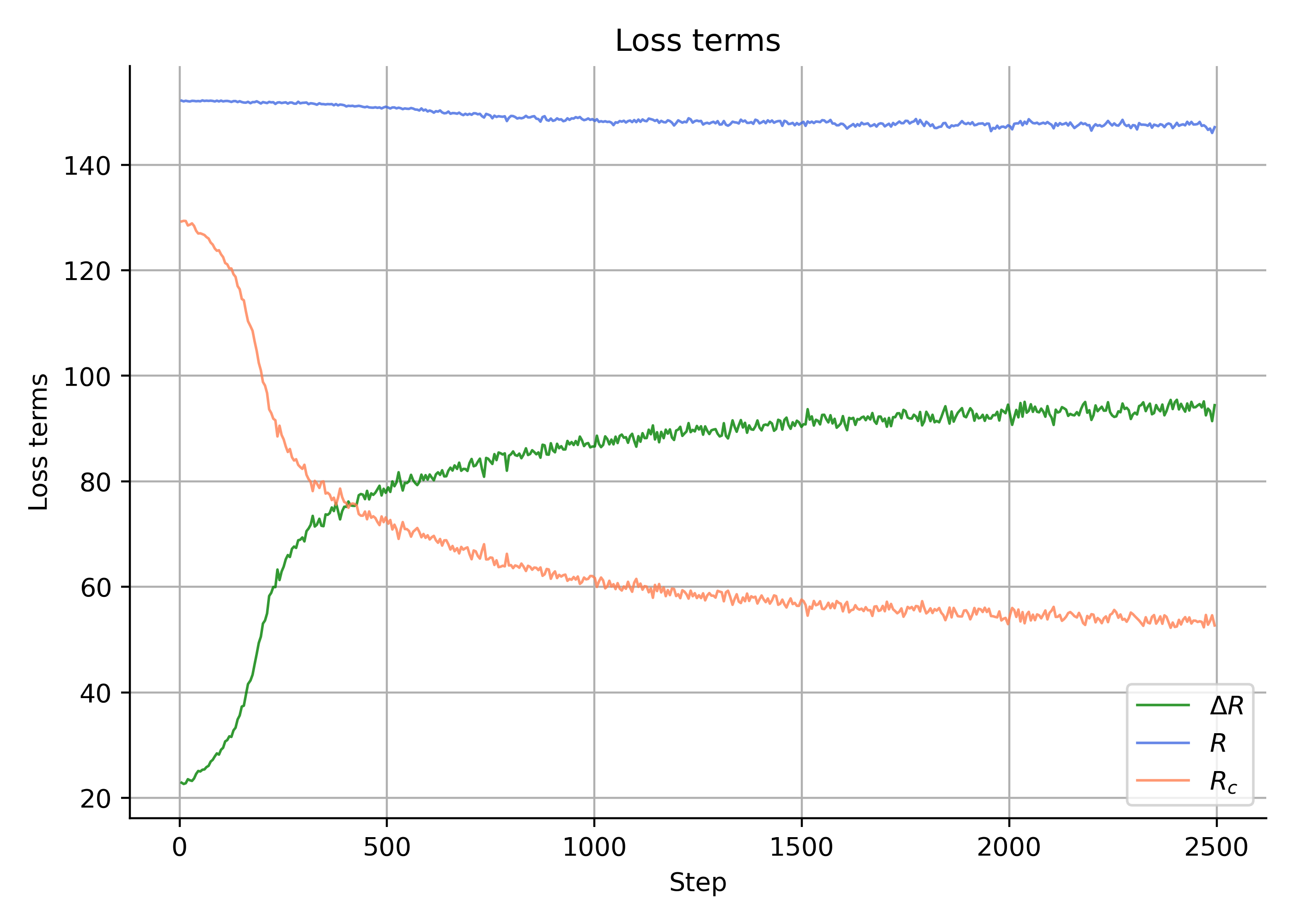}}
        \hfill
      \subfloat[Numerical ranks of all features $\mZ$ and features from each ground-truth cluster $i$, $\{\vz_j: y(j)=i\}$.\label{fig:cifar10-b}]{\includegraphics[width=0.49\linewidth]{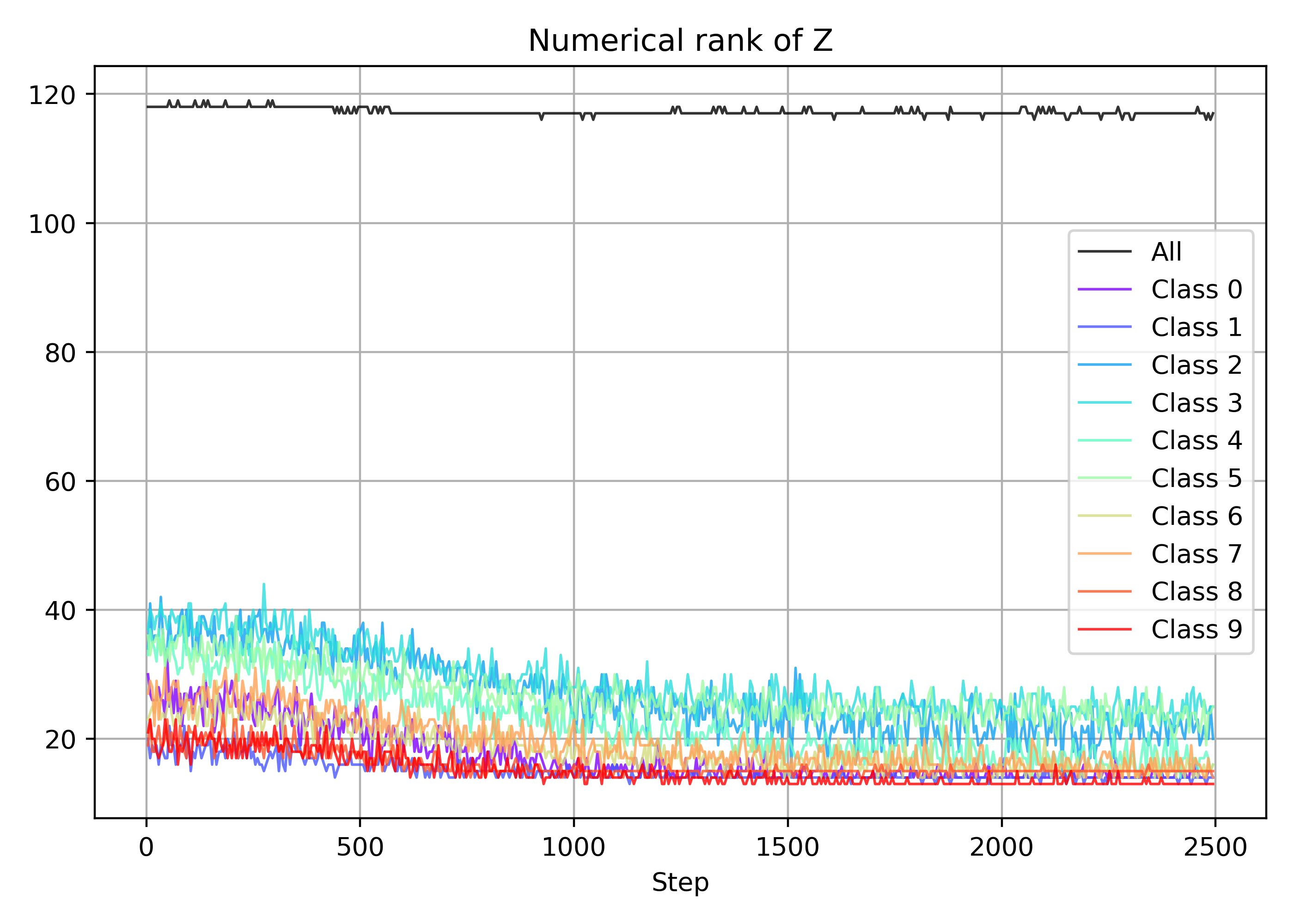}}
      \caption{Coding rates (as loss terms in \mours{} (\ref{eq:mcr2-clustering})) and numerical ranks (\S \ref{sec:exp-sc}) of the features learned by \mours{} on CIFAR-10 as epoch varies.}
      \label{fig:cifar10-lines}
      \end{minipage}
      \hfill
    \begin{minipage}[b]{0.32\textwidth}
      \centering
       \includegraphics[width=\linewidth]{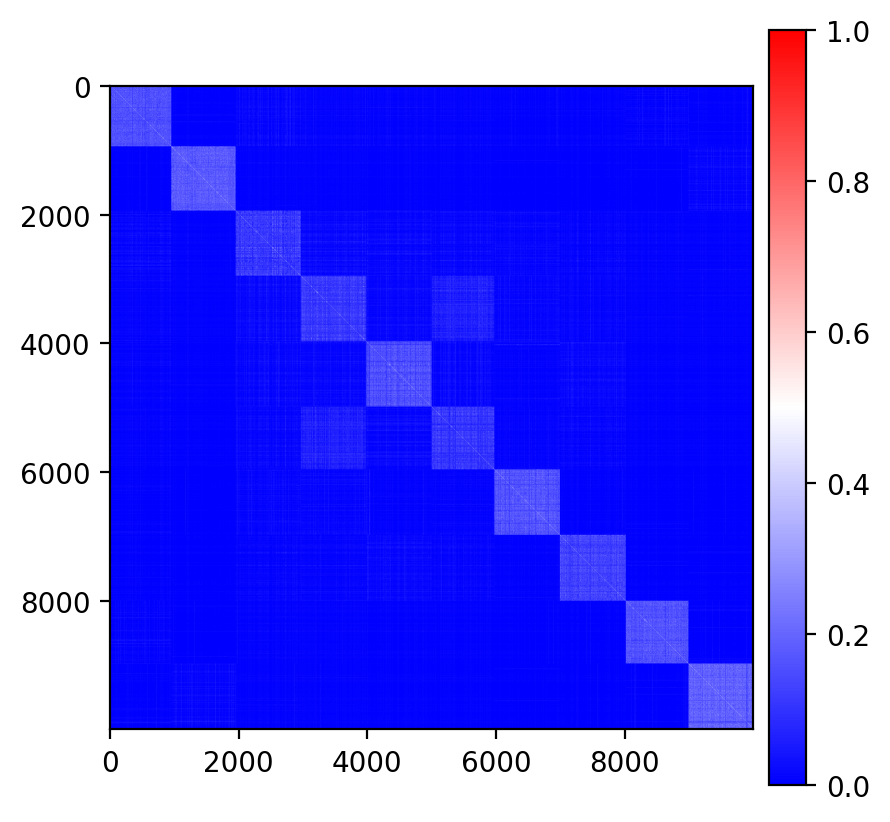}
      \caption{Cosine similarity $|\mZ_{\text{\mours}}^\top \mZ_{\text{\mours}}|$ of the features $\mZ_{\text{\mours}}$ learned by \mours{}.}\label{fig:cifar10-ztz}
    \end{minipage}
  \end{figure*}

   \subsection{Comparison with Subspace Clustering} \label{sec:exp-sc}
   To demonstrate the ability of \mours{} to cluster the samples and linearize the manifolds, we conduct experiments on CIFAR-10, which consists of RGB images from $10$ classes such as planes, birds, and deers. As mentioned in \S \ref{sec:intro} subspace clustering methods rely crucially on the assumption that data lie close to a union of linear subspaces, which many real-world dataset may not satisfy. To show that this is the case, we additionally compare the proposed method with subspace clustering methods. As we shall see, applying subspace clustering directly on self-supervised features of CIFAR-10 will yield low clustering accuracy. In contrast, \mours{} is able to \textit{achieve high clustering accuracy}, and moreover, \textit{produce a union-of-orthogonal-subspaces representation} on which subspace clustering methods can also achieve high accuracy. 
   
   \myparagraph{Data} We use the training split of CIFAR-10 containing $50000$ RGB images, each of size $3\times 32\times 32$. We use the augmentation specified in the Appendix
   to perform self-supervised representation learning via \texttt{TCR} (\ref{eq:opt-tcr-sup}) and get $\mZ_{\text{\texttt{TCR}}}$. For a fair comparison, the so-learned $\mZ_{\text{\texttt{TCR}}}$ are used both as initialization for \mours{} (line \ref{alg-line:self-sup-init} of \Cref{alg:mcr2-clustering}), and as the input for subspace clustering methods\footnote{Self-supervised features learned via coding rate measures empirically exhibit some union-of-subspace structure (one subspace per cluster). They have been used for subspace clustering as in, e.g., \cite[\S 3.2]{Yu2020-mx} and \cite[\S 4.2]{Zhang2021-sx}. }. In \mours{}, for each image in each batch we randomly sample $A=2$ augmentations to apply to the image. As an additional comparison, we also run subspace clustering methods on the features $\mZ_{\text{\mours}}$ learned by \mours.
   
   \myparagraph{Methods} We compare with the elastic-net subspace clustering with active-set solver (\texttt{EnSC}, \cite{You2016-ob}) and sparse subspace clustering with orthogonal matching pursuit solver (\texttt{SSC}-\texttt{OMP},~\cite{You2016-na}), using off-the-shelf implementation provided by the authors\footnote{\scriptsize\url{https://github.com/ChongYou/subspace-clustering}.}. We search the parameters of \texttt{EnSC} over $(\gamma,\tau)\in\{1,5,10,50,100\}\times\{0.9,0.95,1\}$ and those of \texttt{SSC} over $(k_{\mathrm{max}},\epsilon)\in\{3,5,10,20\}\times\{10^{-4},10^{-5},10^{-6},10^{-7}\}$, and report the run with the highest clustering accuracy for each method. 
   We summarize detailed parameters for \mours{} in the Appendix.

   \begin{table}[t]
       \centering
       \small
       \setlength{\tabcolsep}{15.5pt}
       \caption{Clustering accuracy and normalized mutual information of subspace clustering (\texttt{EnSC}, \texttt{SSC}-\texttt{OMP}) and proposed manifold linearizing and clustering (\mours{}). $\mX$ contains $6\cdot 10^{4}$ images from $10$ classes of CIFAR-10, $\mZ_{\text{\texttt{TCR}}}$ are features learned via self-supervised learning \texttt{TCR}, $\mZ_{\text{\mours}}$ are features learned by \mours.} 
       \label{tab:sc}
       \begin{tabular}{@{}lllrr@{}}
       \toprule
       Method &  & Input Data & ACC & NMI \\ \midrule
       \multirow{2}{*}{\texttt{EnSC}} &  & $\mZ_{\text{\texttt{TCR}}}$ & 72.2 & 67.9 \\
        &  & $\mZ_{\text{\mours}}$ & 81.5 & \textbf{79.2} \\ \midrule
       \multirow{2}{*}{\texttt{SSC}-\texttt{OMP}} &  & $\mZ_{\text{\texttt{TCR}}}$ & 67.8 & 64.5 \\
        &  & $\mZ_{\text{\mours}}$  & 78.4 & 76.3 \\ \midrule
       \mours &  & $\mX$ & \textbf{86.3} & 78.3 \\ \bottomrule
       \end{tabular}
   \end{table}

   \myparagraph{Results} \Cref{fig:cifar10-lines} reports the coding rates (as loss terms in (\ref{eq:mcr2-clustering}) and numerical ranks of features learned by \mours{} as epoch varies. As a first note, the coding rate $R$ of all features (the blue curve in \ref{fig:cifar10-a}) decreases only slightly as epoch goes, indicating that the overall representation is diverse in the feature space. Indeed, the numerical rank of all features (the dark curve in \Cref{fig:cifar10-b}) stays $118$ which is close to the dimension $128$ of the feature space. This is in sharp contrast to the deep subspace clustering methods where all the features collapse to a one-dimensional subspace~\cite{Haeffele2020-kj}. Moreover, as the coding rate $R_c$ of clustered features (the orange curve in \Cref{fig:cifar10-a}) goes down, the numerical ranks of features from each ground-truth cluster decrease. For instance, the representation from true cluster $3$ has a numerical rank of $37$ in the first step and $24$ in the last step. This implies that most representation gets linearized better and clustered more accurately, even though the \mours{} objective \eqref{eq:mcr2-clustering} is unsupervised, i.e., it does not use ground-truth labels. Last but not least, note that the features within each ground-truth cluster spread well in a low-dimensional subspace, e.g., the numerical ranks for the true clusters at the last step are within $[13, 23]$. This achieves the desired within-cluster diverse property (\S \ref{sec:intro}), as opposed to the neural collapse phenomenon that appears with the cross-entropy loss.

   Now we compare \mours{} with subspace clustering methods. \Cref{tab:sc} reports clustering accuracy and normalized mutual information for applying \texttt{EnSC}, \texttt{SSC}-\texttt{OMP} on self-supervised features $\mZ_{\text{\texttt{TCR}}}$, features $\mZ_{\text{\mours}}$ learned by \mours, and applying \mours{} on $\mX$, where $\mX$ is $6\cdot 10^{4}$ images from $10$ classes of CIFAR-10. In addition, we plot the cosine similarity of the features learned by \mours{} in \Cref{fig:cifar10-ztz}.
   Remarkably, the highest clustering accuracy is $86.3\%$ achieved by \mours{} on $\mX$, which surpasses \texttt{EnSC} ($72.2\%$) and \texttt{SSC}-\texttt{OMP} ($67.8\%$) on $\mZ_{\text{\texttt{TCR}}}$ by a large margin, even though $\mZ_{\text{\texttt{TCR}}}$ is used both as initialization for \mours{} and input for \texttt{EnSC} and \texttt{SSC}-\texttt{OMP}. Interestingly, using instead the features $\mZ_{\text{\mours}}$ learned by \mours{}, the clustering performance of \texttt{EnSC} and \texttt{SSC}-\texttt{OMP} increases and even becomes comparable to \mours{}, e.g., \texttt{EnSC} achieves $79.2\%$ normalized mutual information compared to $78.3\%$ of \mours{}. This suggests that $\mZ_{\text{\mours}}$ has a union-of-subspace structure that can be utilized by subspace clustering. Indeed, as seen in \Cref{fig:cifar10-ztz}, features from different clusters tend to have a small similarity, i.e., being orthogonal to each other. This demonstrates the between-cluster discrimination (\S \ref{sec:intro}) as desired.

   \subsection{Comparison with Deep Clustering} \label{sec:exp-deep}
   We further compare the proposed \mours{} with state-of-the-art deep clustering methods on large-scale datasets (CIFAR-10, -20, -100, and TinyImageNet-200). Different than \mours{}, most methods reported (all except \texttt{NMCE} as discussed in \S \ref{sec:mcr2-unsup-formulation}) do \textit{not} aim to learn a union-of-orthogonal-subspaces representation. Be that as it may, \mours{} achieves comparable or better clustering accuracy than state-of-the-art methods using less running time, even when the dataset presents many or imbalanced clusters.

    \myparagraph{Datasets}
   Beyond CIFAR-10 (\S \ref{sec:exp-sc}), we further use CIFAR-20, CIFAR-100 and TinyImageNet-200 to evaluate the performance of our method. Both CIFAR-100 and CIFAR-20 contain the same $50000$ train images and $10000$ test images with size $32\times32\times3$, while the former are split into $100$ clusters and the latter $20$ super clusters. Finally, TinyImageNet contains $100000$ train images and $10000$ test images with size $64\times64\times3$ split into $200$ clusters.

   \myparagraph{Baseline Methods} We include clustering accuracy and normalized mutual information reported by \texttt{SCAN} \cite{Van_Gansbeke2020-eo}, \texttt{GCC} \cite{Zhong2021-gl}, \texttt{NNM} \cite{Dang2021-no}, \texttt{IMC} \cite{Ntelemis2021-gz}, \texttt{NMCE} \cite{Li2022-vq}, \texttt{SPICE} \cite{Niu2022-iq} on aforementioned datasets whenever applicable. In addition, to \textit{compare the running time} as well as to have more baseline methods when there are \textit{many clusters}, we conduct experiments with \mours{}, \texttt{SCAN} \cite{Van_Gansbeke2020-eo}, and \texttt{IMC} \cite{Ntelemis2021-gz} on CIFAR-100. Training details are left to \S \ref{sec:details-real-exp}. 
   For a fair comparison, all methods reported use ResNet-18 as the backbone, commonly adopted by the literature. Note that each method chooses one or more pre-training that best fits its objective, such as \texttt{SimCLR} \cite{chen2020simple}, \texttt{SwAV} \cite{Caron2020-el}, \texttt{TCR} \cite{Li2022-vq}, \texttt{MoCoV2} \cite{Chen2020-ws}. Hence, for clarity, we indicate the pre-training used after the method, e.g., \texttt{SCAN}-\texttt{SimCLR} means \texttt{SCAN} method initialized with \texttt{SimCLR}.

   \begin{table}[t]
    \centering
    \caption{Clustering accuracy and normalized mutual information of different methods on CIFAR-10 and CIFAR-20. For a fair comparison, all methods use ResNet-18 as backbone. 
    }
    \label{tab:exp-cifar10-20}
    \resizebox{\columnwidth}{!}{%
    \begin{tabular}{@{}llcccc@{}}
    \toprule
    Method vs.  &  & \multicolumn{2}{c}{CIFAR-10}  & \multicolumn{2}{c}{CIFAR-20}  \\
    Dataset \& Metric             &           & ACC  & NMI  & ACC  & NMI  \\ \midrule
    \texttt{SCAN}-\texttt{SimCLR}$_{\text{ (ECCV '20)}}$  & & .876 & .787 & .468 & .459 \\
    \texttt{GCC}-\texttt{SimCLR}$_{\text{ (ICCV '21)}}$          & & .856 & .764 & .472 & .472 \\
    \texttt{NNM}-\texttt{SimCLR}$_{\text{ (CVPR '21)}}$ & & .843 & .748 & .477 & .484 \\
    \texttt{IMC}-\texttt{SwAV}$_{\text{ (KBS '22)}}$      & & \textbf{.891} & \textbf{.811} & \textbf{.490} & \textbf{.503} \\ \midrule
    \texttt{NMCE}-\texttt{TCR}$_{\text{ (Arxiv '22)}}$    & & .830 & .761 & .437 & .488 \\
    \mours{}-\texttt{TCR}$_{\text{ (Ours)}}$         & & \textbf{.863} & \textbf{.783} & \textbf{.522} & \textbf{.546} \\ \midrule
    \texttt{SCAN}-\texttt{MoCoV2}$_{\text{ (ECCV '20)}}$  & & .874 & .786 & .455 & .472 \\
    \texttt{SPICE}-\texttt{MoCoV2}$_{\text{ (TIP '22)}}$  & & .918 & .850 & .535 & .565 \\
    \mours{}-\texttt{MoCoV2}$_{\text{ (Ours)}}$      & & \textbf{.922} & \textbf{.855} & \textbf{.583} & \textbf{.596} \\ \bottomrule
    \end{tabular}%
    }
\end{table}
\begin{table}[t]
  \centering
  \caption{Clustering accuracy and normalized mutual information of different methods on CIFAR-100 and TinyImageNet-200. For a fair comparison, all methods use ResNet-18 as backbone. 
   }
  \label{tab:exp-cifar100-timgnet200}
  
  \resizebox{\columnwidth}{!}{%
  \begin{tabular}{@{}llcccc@{}}
  \toprule
  Method vs.  &  & \multicolumn{2}{c}{CIFAR-100} & \multicolumn{2}{c}{TinyImageNet-200} \\ 
  
  Dataset \& Metric &  & ACC & NMI & ACC & NMI \\ 
  
  \midrule
  \texttt{SCAN}-\texttt{SimCLR}$_\text{ (ECCV '20)}$ &  & 34.3 & 55.7 & - & - \\
  \texttt{GCC}-\texttt{SimCLR}$_{\text{ (ICCV '21)}}$ &  & - & - & 13.8 & 34.7 \\
  \texttt{IMC}-\texttt{SwAV}$_{\text{ (KBS '22)}}$ &  & 43.9 & 58.3 & 28.2 & 52.6 \\
  \texttt{SPICE}-\texttt{MoCoV2}$_{\text{ (TIP '22)}}$ &  & - & - & 30.5 & 44.9 \\
  \mours-\texttt{TCR}$_{\text{ (Ours)}}$ &  & \textbf{49.4} & \textbf{68.3} & \textbf{33.5} & \textbf{67.5} \\
  \bottomrule
  \end{tabular}%
  }
  \end{table}

    \begin{table}[t]
    \centering
    \setlength{\tabcolsep}{5.5pt}
        \caption{Running time in minutes and clustering accuracy on CIFAR-100. For a fair comparison, all methods use ResNet-18 as backbone.}
        \label{tab:Runnning_time}
        \resizebox{\columnwidth}{!}{%
        \begin{tabular}{@{}llccccc@{}}
        \toprule
        Method vs.  &  & \multicolumn{4}{c}{Running Time} & ACC \\ 
        Metric \& Stage &  & I & II& III & Total  \\ 
        \midrule
        
        \texttt{SCAN}-\texttt{SimCLR}$_\text{ (ECCV '20)}$ &  & 308.3 & 33.3 & 54.7 & 396.3 &34.3 \\
        
        \texttt{IMC}-\texttt{SwAV}$_{\text{ (KBS '22)}}$ &  & 529.4 & - & - & 529.4 &43.9\\
        
        \mours-\texttt{TCR}$_{\text{ (Ours)}}$ &  & 266.7 & 17.7 & -  & \textbf{284.4} & \textbf{49.4}  \\
        
         \bottomrule
        \end{tabular}%
        }
  \end{table}

  \myparagraph{Results on CIFAR-10, -20} Table \ref{tab:exp-cifar10-20} presents clustering accuracy and normalized mutual information of various methods. Overall, \mours{} is \textit{the most accurate}, among methods using either the same pre-training (middle and bottom rows) or any pre-training. To begin with, using \texttt{TCR} as pre-training, \mours{} achieves $86.3\%$ and $52.2\%$ clustering accuracies on CIFAR-10 and -20, which are $3.3$ and $8.5$ percentage points higher respectively than those achieved by \texttt{NMCE}\footnote{
  For a further comparison between \mours{} and \texttt{NMCE}, see the end of \S \ref{sec:mcr2-unsup-formulation} for conceptual and algorithmic differences, and \S \ref{sec:appendix-nmce} for an empirical study on the stability against random seeds.}. Remarkably, when \mours{} is initialized with \texttt{MoCoV2} pre-training, it yields even higher clustering accuracies of $92.2\%$ on CIFAR-10 and $58.3\%$ on CIFAR-20, \textit{surpassing previous state-of-the-art methods} \texttt{SPICE}-\texttt{MoCoV2} ($91.8\%$, $53.5\%$) and \texttt{IMC}-\texttt{SwAV} ($89.1\%$, $49.0\%$). Interestingly, while the clustering performance of \mours{}-\texttt{TCR} is competitive on CIFAR-20, it is less so on CIFAR-10. After investigation, we find that the learned clusters appear \textit{semantically meaningful}, even though they do not agree with the ground-truth labels of CIFAR-10 used for evaluation; we leave the details to \S \ref{sec:semantic}.

    \myparagraph{Results on CIFAR-100, TinyImageNet-200} We report clustering accuracy and normalized mutual information on both datasets in Table \ref{tab:exp-cifar100-timgnet200}, and further show running time on CIFAR-100 in Table \ref{tab:Runnning_time}. Notably, \mours{} outperforms \texttt{SCAN} and \texttt{IMC}-\texttt{SwAV} on CIFAR-100 and TinyImageNet-200 by a large margin, while using \textit{lower running time}: E.g., on CIFAR-100, \mours{} yields an accuracy of $49.4\%$ in $291$ minutes, whereas \texttt{IMC}-\texttt{SwAV} has $43.9\%$ using $529$ minutes, and \texttt{SCAN} has $34.3\%$ in $396$ minutes.

  \begin{table}[t]
    \centering
    \caption{Clustering accuracy on imbalanced datasets: (a) Imb-CIFAR-10, (b) Imb-CIFAR-100. For a fair comparison, all methods use ResNet-18 as backbone.  }
    \label{tab:Imbalance_data}
    \begin{tabular}{@{}lccc@{}}
    \toprule
    Method / Dataset &  & (a) & (b) \\
    \midrule
    \texttt{SCAN}$_\text{ (ECCV '20)}$ &  & 62.9 & 31.1 \\
    \texttt{IMC}-\texttt{SwAV}$_{\text{ (KBS '22)}}$ &  & 65.7 & 38.2  \\
    \mours$_{\text{ (Ours)}}$ &  & \textbf{80.0} & \textbf{46.1} \\
    \bottomrule
    \end{tabular}%
  \end{table}

   \myparagraph{Results on Imbalanced Clusters}
         Note that for CIFAR-10 or CIFAR-100 each cluster contains approximately the same number of samples. On the other hand, natural images are typically imbalanced, i.e., the clusters have unequal number of samples. To mimic this setting, we take a naive approach to construct the following imbalanced datasets. For the $10$ clusters of CIFAR-10, we remove half of the samples from odd-numbered clusters (i.e., clusters $1,3,\dots,9$) from both the training and test split. We refer to the reduced dataset Imb-CIFAR-10. Likewise we construct Imb-CIFAR-100. We run two state-of-the-art methods \texttt{IMC}-\texttt{SwAV} and \texttt{SCAN} as well as the proposed \mours{} on Imb-CIFAR-10 and Imb-CIFAR-100.

   \Cref{tab:Imbalance_data} shows clustering accuracy on the imbalanced datasets Imb-CIFAR-10 and Imb-CIFAR-100. As a first observation, the clustering accuracy of all methods is lower on the imbalanced datasets than on the balanced counterparts, as expected. Notably, \mours{} suffers from the least performance drop, e.g., when moving from CIFAR-10 to Imb-CIFAR-10 the accuracy of \mours{} drops from $86\%$ to $80\%$, whereas that of \texttt{SCAN} and \texttt{IMC}-\texttt{SwAV} decreases from above $87\%$ to below $66\%$.

   \section{Conclusion}
   This paper studies the problem of simultaneously clustering and learning an union-of-orthogonal-subspaces representation for data, when data lies close to a union of low-dimensional manifolds. To address the problem we propose an objective based on {\em maximal coding rate reduction} and {\em doubly stochastic} membership inspired by the state-of-the-art subspace clustering results. We provide an efficient and effective  parameterization of the membership variables as well as a meta-algorithm to optimize the representation and membership jointly. We further conduct experiments on datasets with larger number of clusters and imbalanced clusters and show that the proposed method achieves state-of-the-art performance. We believe that our work provides a general and unified framework for unsupervised learning of structured representations for multi-modal data.

   \medskip \noindent
\textbf{Acknowledgements}
This work was partially supported by ONR grant N00014-22-1-2102, the joint Simons Foundation-NSF DMS grant 2031899, a research grant from TBSI, NSF grant 1704458, the Northrop Grumman Mission Systems Research in Applications for Learning Machines (REALM) initiative, and NSF graduate fellowship DGE2139757.

   {\small
   \bibliographystyle{ieee_fullname}
   \bibliography{references,td_paperpile}
   }
   
   \clearpage
   \appendix
   \onecolumn
   
   \begin{figure*}
     \centering
     \begin{subfigure}{0.32\textwidth}
       \includegraphics[width=\linewidth,trim={2cm 2cm 2cm 2cm},clip]{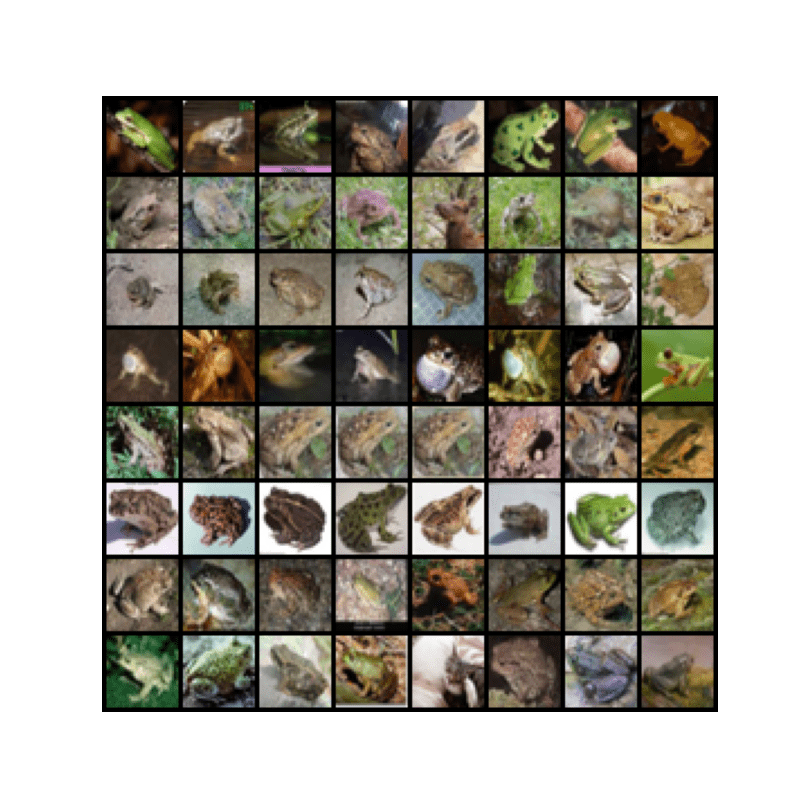}
       \caption{Learned cluster 1}
       \label{fig:cifar10-1}
     \end{subfigure}
     \hfill
     \begin{subfigure}{0.32\textwidth}
       \includegraphics[width=\linewidth,trim={2cm 2cm 2cm 2cm},clip]{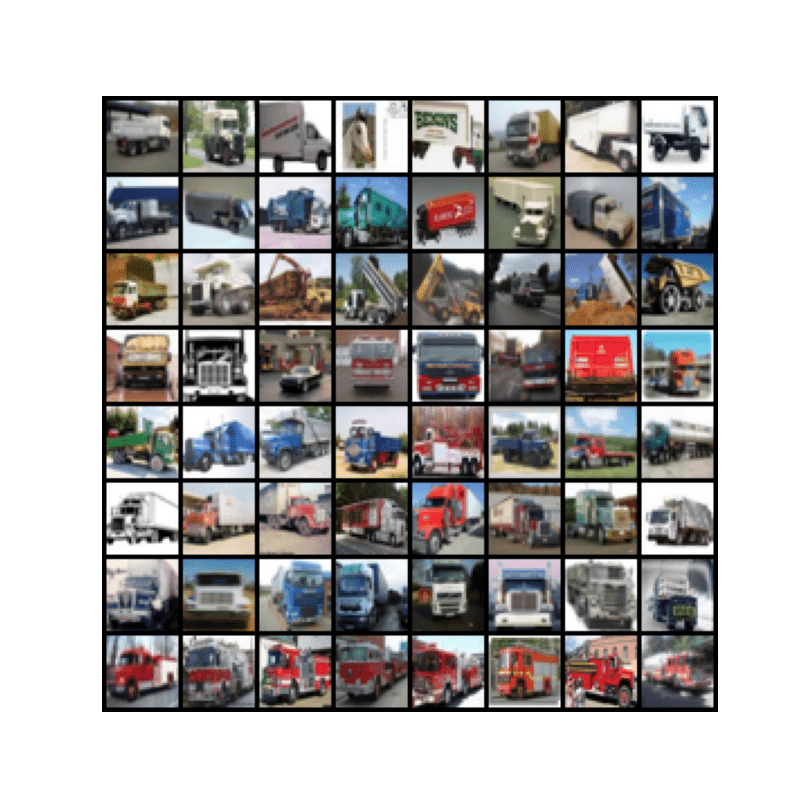}
       \caption{Learned cluster 2}
       \label{fig:cifar10-2}
     \end{subfigure}
     \hfill
     \begin{subfigure}{0.32\textwidth}
       \includegraphics[width=\linewidth,trim={2cm 2cm 2cm 2cm},clip]{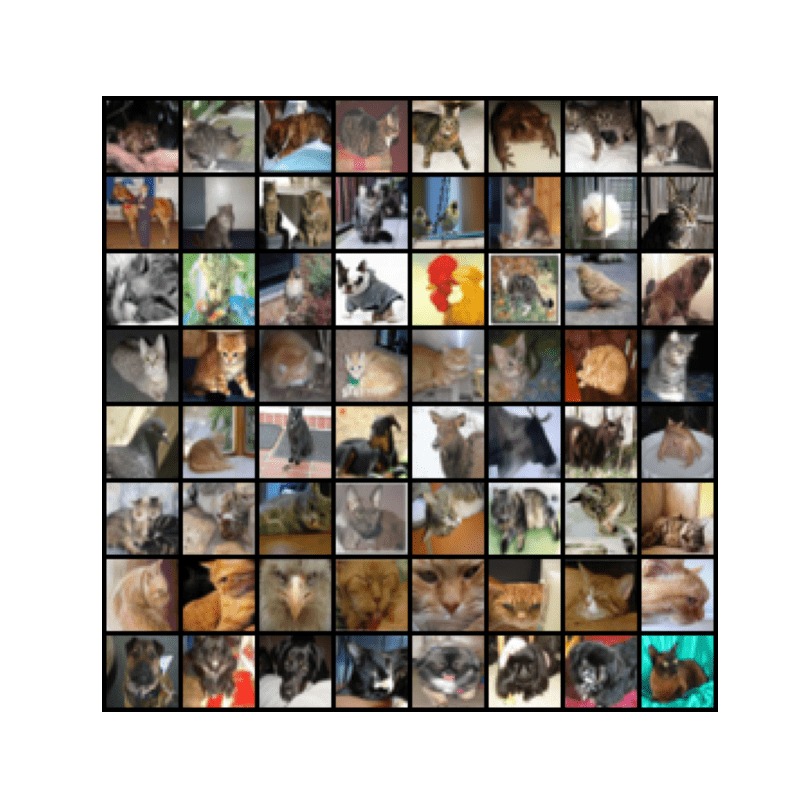}
       \caption{Learned cluster 3}
       \label{fig:cifar10-3}
     \end{subfigure}
     \hfill
     \begin{subfigure}{0.32\textwidth}
       \includegraphics[width=\linewidth,trim={2cm 2cm 2cm 2cm},clip]{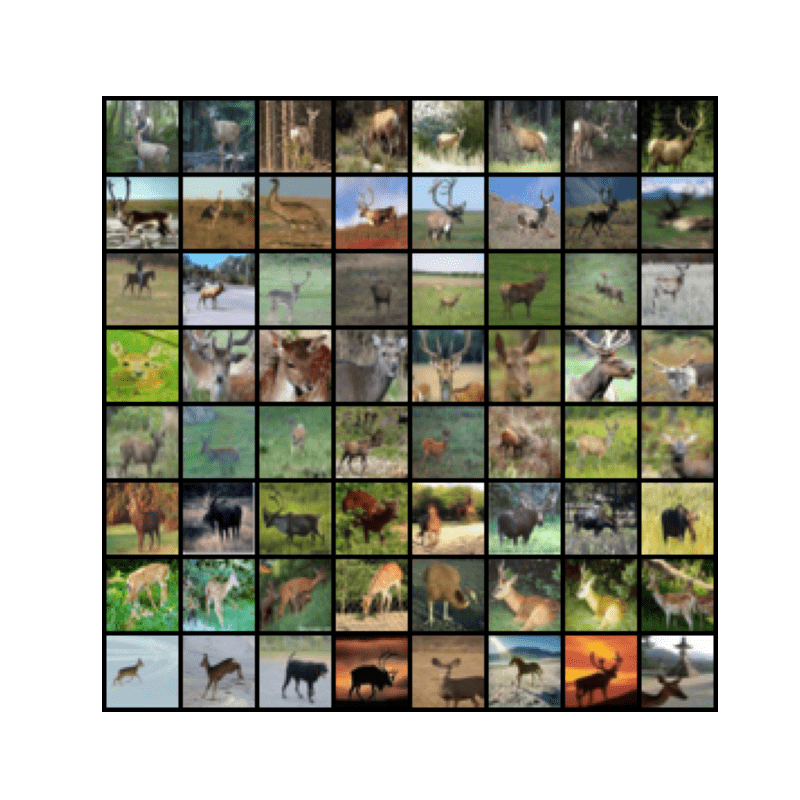}
       \caption{Learned cluster 4}
       \label{fig:cifar10-4}
     \end{subfigure}
     \hfill
     \begin{subfigure}{0.32\textwidth}
       \includegraphics[width=\linewidth,trim={2cm 2cm 2cm 2cm},clip]{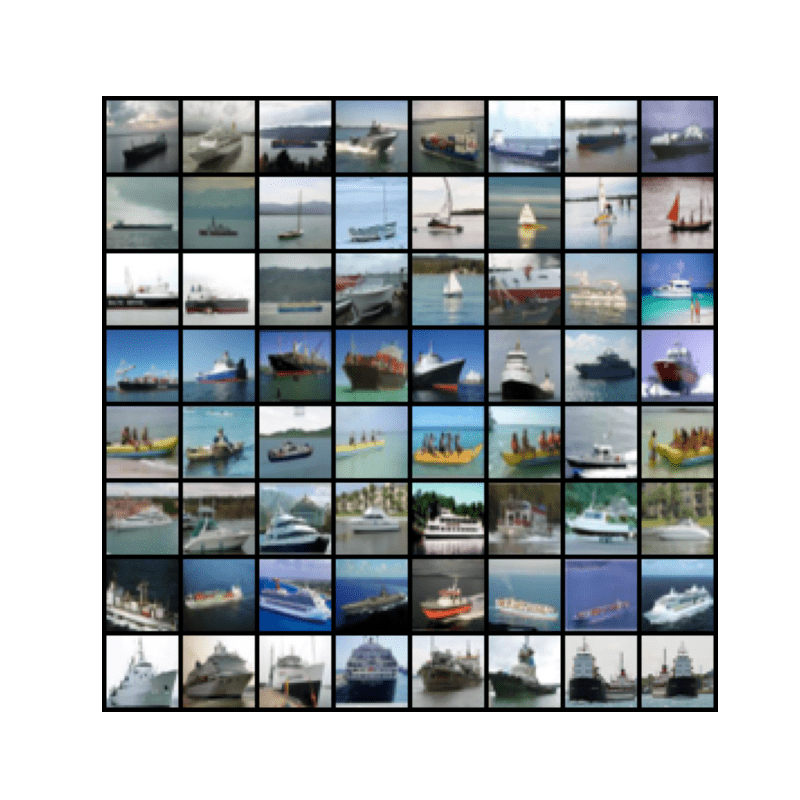}
       \caption{Learned cluster 5}
       \label{fig:cifar10-5}
     \end{subfigure}
     \hfill
     \begin{subfigure}{0.32\textwidth}
       \includegraphics[width=\linewidth,trim={2cm 2cm 2cm 2cm},clip]{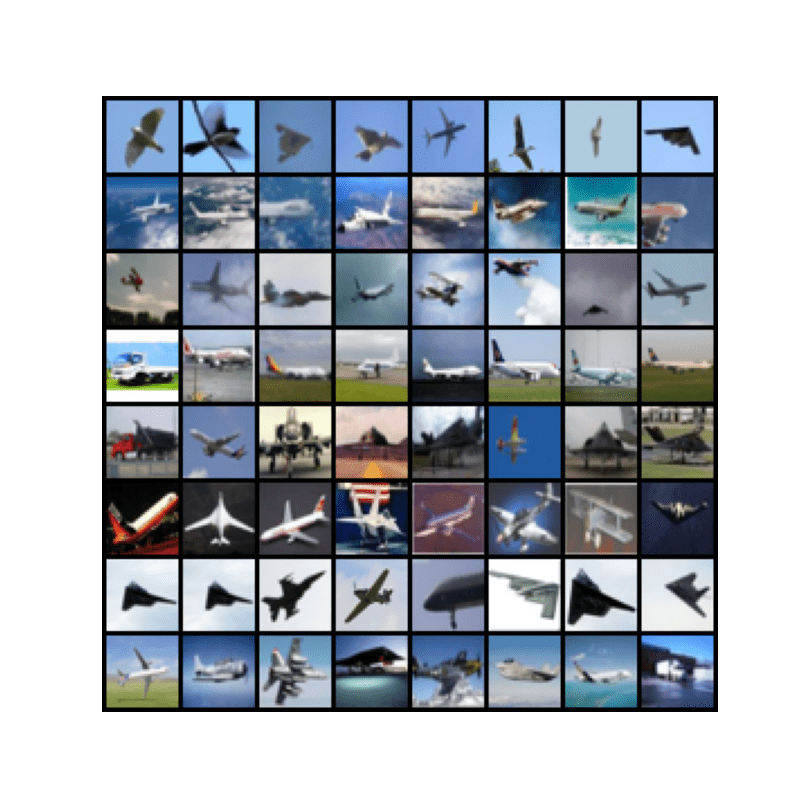}
       \caption{Learned cluster 6}
       \label{fig:cifar10-6}
     \end{subfigure}
     \hfill
     \begin{subfigure}{0.32\textwidth}
       \includegraphics[width=\linewidth,trim={2cm 2cm 2cm 2cm},clip]{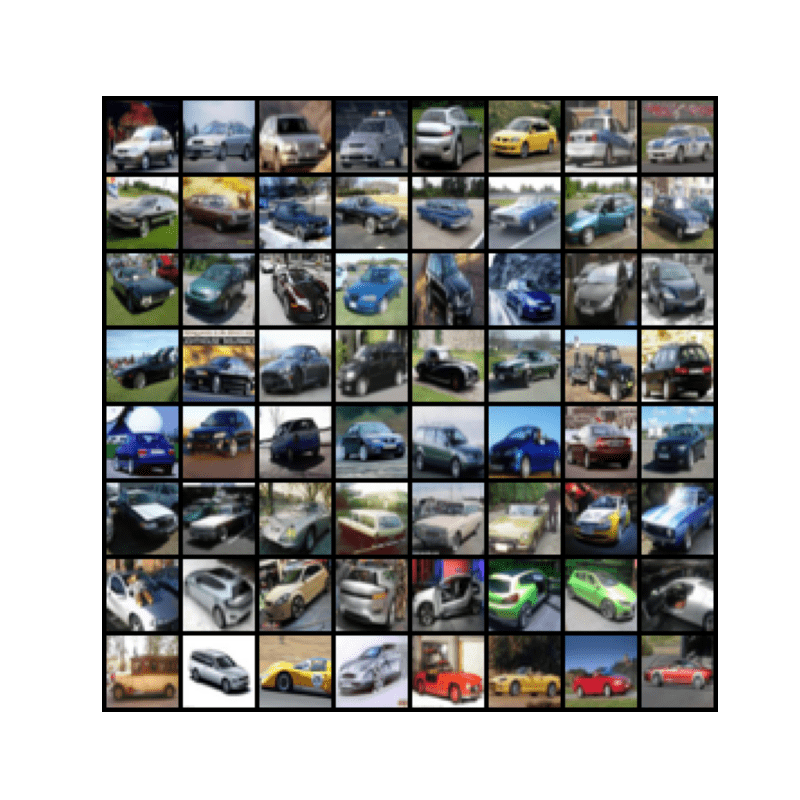}
       \caption{Learned cluster 7}
       \label{fig:cifar10-7}
     \end{subfigure}
     \hfill
     \begin{subfigure}{0.32\textwidth}
       \includegraphics[width=\linewidth,trim={2cm 2cm 2cm 2cm},clip]{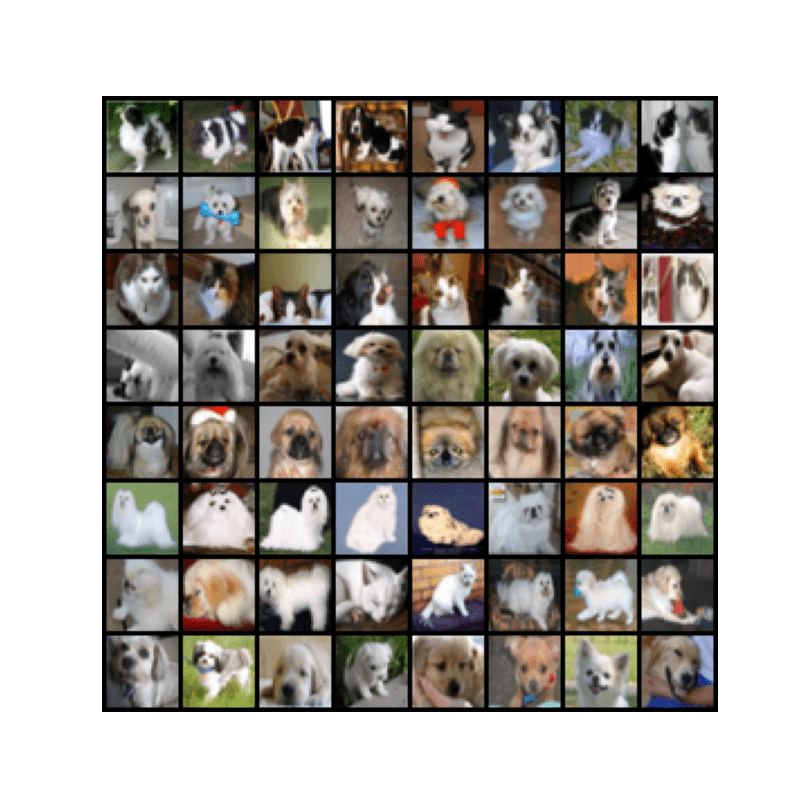}
       \caption{Learned cluster 8}
       \label{fig:cifar10-8}
     \end{subfigure}
     \hfill
     \begin{subfigure}{0.32\textwidth}
       \includegraphics[width=\linewidth,trim={2cm 2cm 2cm 2cm},clip]{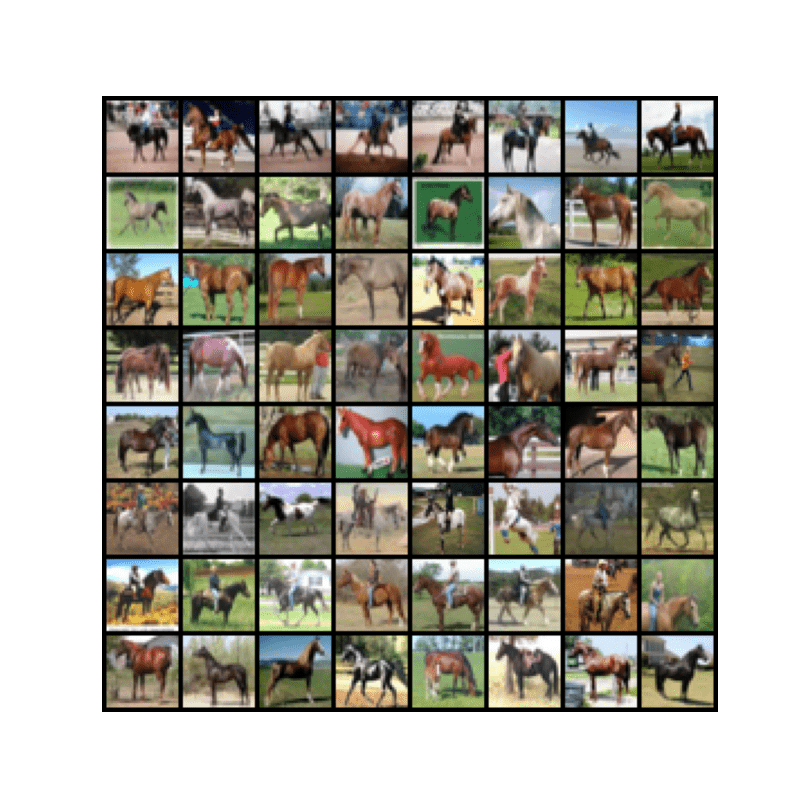}
       \caption{Learned cluster 9}
       \label{fig:cifar10-9}
     \end{subfigure}
     \caption{Images along the principal components (see \S \ref{sec:semantic}) of features from each cluster on CIFAR-10, where features and clusters are learned by \mours{}. For the sake of space, only $9$ clusters are shown. }
     \label{fig:cifar10-pc}
   \end{figure*}
   
   \clearpage
   
   \section{Semantic Interpretability and Failure Case Analysis on CIFAR-10 }\label{sec:semantic}
   Recall that \mours{} clusters images and learns a representation for them where images of each cluster lie close to a low-dimensional subspace. A natural question that arises is what are the \textit{different directions within each subspace}? Below we give some interesting visualization. Specifically, after \mours{} is trained, we take the (learned) features from each (learned) cluster, and apply Principal Component Analysis to them to obtain the first $8$ \textit{principal components} (for a review see, e.g., \cite[\S 2.1]{Vidal2016-zp}). Recall that principal components are mutually orthogonal, indicating uncorrelated directions in the subspace (of each cluster). To visualize these principal components, we show the images from each cluster whose features are the closest to the principal components.

    \Cref{fig:cifar10-pc} reports the images along the principal components of the $10$ clusters using the representation and clusters learned by \mours{}, where each sub-figure corresponds to a cluster and each row a principal component. Interestingly, the rows of images appear to exhibit some semantic `concepts': in \Cref{fig:cifar10-2}, row $1$ and $8$ are respectively white and red trucks, while row $3$ are the trucks that ship sand or mud; row $1$ of \Cref{fig:cifar10-4} are deers with trees as background. This seems to suggest that the learned representation preserves distance within each cluster, i.e., images that are close/far in semantic meaning will be close/far in the feature space, as desired in \S \ref{sec:intro}. 

    This visualization further allows us to investigate why on CIFAR-10 \mours{} has a clustering accuracy slightly lower than some state-of-the-art methods (\Cref{tab:exp-cifar10-20}). We observe that the main  clusters samples are in clusters $3$ and $8$: 
    e.g., rows $1$ and $3$ of \Cref{fig:cifar10-8} are cats while all other rows in this cluster are dogs. On the other hand, one may argue that \Cref{fig:cifar10-8} is a cluster of pets of lighter colors, and \Cref{fig:cifar10-3} a cluster of pets of darker colors. They could be semantically meaningful clusters despite not aligning with the ground-truth labels. We believe it would be an interesting future work to use \mours{} to discover new semantics that are not present in the given labels.

\section{Details on Experiment Settings}
   \subsection{Synthetic Union-of-Manifold Data} \label{sec:simulations}
   We perform simulations
   to visualize the properties of the proposed manifold learning and clustering method. As seen in \Cref{fig:sim1-a}, we generate data $\mX$ from two manifolds on the sphere $\sS^2$, each consisting of $200$ samples. The points from the first manifold (green) take the form 
   {\small
   \begin{align}
       \vx_i=\begin{bmatrix}
           \cos\big(A\sin(\omega\phi_i)\big)\cos\phi_i \\
           \cos\big(A\sin(\omega\phi_i)\big)\sin\phi_i \\
           \sin\big(A\sin(\omega\phi_i)\big)
       \end{bmatrix}
       + \bepsilon_i, 
   \end{align}
   }
   where $A=0.2$ and $\omega=5$ sets the curvature of the manifold, $\bepsilon_i\sim \gN(\vzero,0.05\mI_3)$ is the additive noise, and we take $\phi_i=\frac{2\pi i}{100}$ for $i=1,\dots,100$ to generate $100$ points. On the other hand, the points from the second manifold (blue) are simply $100$ samples from $\gN([0,0,1]^\top,0.05\mI_3)$. We take the feature dimension $d=3$ to be equal to he input dimension $D=3$. We paramterize both the feature head $f_{\btheta}$ and the cluster head $g_{\btheta}$ 
   to be a simple fully-connected network with $100$ hidden neurons, followed by a Rectified Linear Unit as non-linearity and a projection operator onto the sphere $\sS^{2}$. \Cref{fig:sim1-b,fig:sim1-c,fig:sim1-d} report the features $\mZ$ with random initialization (i.e., before line \ref{alg-line:self-sup-init} of \Cref{alg:mcr2-clustering}), with self-supervised initialization, and at convergence of \mours{}.
   Notably, despite $\mZ$ being noisy and only approximately piece-wise linear, as epoch goes $\mZ$ gradually transform to two linear subspaces: the green points converge to a $2$-dimensional subspace (intersected with $\sS^2$) and the blue points converge to a $1$-dimension subspace.
   
   \subsection{Real-World Datasets} \label{sec:details-real-exp}
   \myparagraph{\mours{}}
   Since \mours{} is based on \mcr{} (\S \ref{sec:mcr2}), we follow \cite{yu2020learning, Li2022-vq} for the choice of batch size and augmentation. For all real-data experiments, we use a batch size of $n_{b}=1024$ and \texttt{VICReg} \cite{Bardes2022-gc} augmentation (Augmentation \ref{aug:vicreg}). $A=2$ augmentations are used, while not using any leads to worse clustering accuracy (\S \ref{sec:appendix-aug}). Doubling $n_b$ or $A$ improves accuracy by $\leq1\%$ on CIFAR10-10 with the cost of increased running time.
   In self-supervised initialization of $\mZ$ (line \ref{alg-line:self-sup-init} of \Cref{alg:mcr2-clustering}), we used \texttt{TCR} (see \eqref{eq:opt-tcr-sup} or \cite{Li2022-vq}) or otherwise \texttt{MoCoV2} \cite{Chen2020-ws}. To train \texttt{TCR}, we use the precision (\S \ref{sec:mcr2}) parameter $\epsilon^2=0.2$, a LARS optimizer \cite{You2017-xl} (as is also done in \cite{Chen2020-zj,Li2022-vq}) with a learning rate of $0.3$ and trained \mours{} for $1000$ epochs. For \texttt{MoCoV2}, we use off-the-shelf pre-trained models\footnote{\scriptsize\url{https://github.com/vturrisi/solo-learn/tree/d27c7130d19035c0ba0af8f90217e78d8ebe7f48}.}. On the other hand, in the training of \mours{} objective \eqref{eq:mcr2-clustering}, we use $\epsilon^2=0.1$ 
   and $\eta=0.175$ for the entropy regularization in the Sinkhorn projection \cite{Eisenberger2022-oo} layer $P_{\Omega, \eta}(\cdot)$. We fix the backbone and for each batch, we perform one update for parameters in the feature head $\mZ$ and one update for parameters in the cluster head $\mC$. For each head we use one SGD optimizer \cite{Robbins1951-jb}  with a learning rate of $10^{-2}$, momentum of $0.9$, and weight decay of $5\cdot 10^{-4}$.

\setcounter{algorithm}{0}
\floatname{algorithm}{Augmentation}

\begin{minipage}{0.45\linewidth}%
\begin{algorithm}[H]
\caption{Augmentations for real datasets}
\label{aug:vicreg}
\begin{algorithmic}
\State \texttt{import torchvision.transforms as t}
\State \texttt{t.Compose([}
\State \quad \texttt{t.RandomResizedCrop(32,scale=(0.04, 1.0)),}
\State \quad\texttt{t.RandomHorizontalFlip(p=0.5),}
\State \quad\texttt{t.RandomGrayscale(p=0.2),}
\State \quad\texttt{t.RandomApply([t.ColorJitter(0.4, 0.4, 0.4, 0.1)], p=0.8),}
\State \quad\texttt{GaussianBlur(p=0.1)}
\State \texttt{])}
\end{algorithmic}
\end{algorithm}

\end{minipage}
\hfill
\begin{minipage}{0.5\linewidth}

\begin{table}[H]
  \caption{Ablation study on the roles of different parts of \Cref{alg:mcr2-clustering} and on using augmentation.}
  \label{tab:ablation-study}
  \centering
  \resizebox{\columnwidth}{!}{%
  \begin{tabular}{@{}ll@{}}
  \toprule
  Ablation Study on CIFAR-10 & \begin{tabular}[c]{@{}l@{}}Clustering\\ Accuracy\end{tabular} \\ \midrule
  Full Algorithm 1 & \multicolumn{1}{c}{\textbf{86.3\%}} \\
    &  \\
  \begin{tabular}[c]{@{}l@{}}Replacing self-supervised initialization (line \ref{alg-line:self-sup-init})\\ with random initialization\end{tabular} & \multicolumn{1}{c}{20.0\%} \\
    &  \\
  \begin{tabular}[c]{@{}l@{}}Replacing updating \mours{} loss \eqref{eq:mcr2-clustering} (lines \ref{alg-line:start-mlc}-\ref{alg-line:end-mlc})\\ with subspace clustering (\texttt{EnSC})\end{tabular} & \multicolumn{1}{c}{73.4\%} \\
    &  \\
  Not using augmentation in updating \eqref{eq:mcr2-clustering} (lines \ref{alg-line:start-mlc}-\ref{alg-line:end-mlc}) & \multicolumn{1}{c}{80.0\%} \\ \bottomrule
  \end{tabular}%
  }
  \end{table}
\vspace{0.2cm}
\end{minipage}

   \myparagraph{\texttt{SCAN} and \texttt{IMC}-\texttt{SwAV}} Recall that we conduct experiments on CIFAR-100, Imb-CIFAR-10, and Imb-CIFAR-100 with \texttt{SCAN} \cite{Van_Gansbeke2020-eo}, \texttt{IMC}-\texttt{SwAV} \cite{Ntelemis2021-gz} and \mours{}, and report clustering and running time in \Cref{tab:Runnning_time,tab:Imbalance_data}. We use off-the-shelf implementation\footnote{\scriptsize\url{https://github.com/wvangansbeke/Unsupervised-Classification}, \url{https://github.com/foiv0s/IMC-SwAV-pub}} provided by the authors. For a fair comparison, \texttt{SCAN}, \texttt{IMC}-\texttt{SwAV} and \mours{} all use ResNet-18 as the backbone. Finally, the hyper-parameters of \texttt{SCAN} and \texttt{IMC}-\texttt{SwAV} are set to be the ones optimally chosen for CIFAR-10 and CIFAR-100 respectively provided by the authors.

   \section{Role and Ablation Study of Augmentation} \label{sec:appendix-aug}
   Recall that data augmentation was used both in the self-supervised initialization
    (line \ref{alg-line:self-sup-init} of \Cref{alg:mcr2-clustering}, see `Initializing $\mZ$' in \S \ref{sec:algo}) and in updating the \mours{} objective  (lines \ref{alg-line:start-mlc}-\ref{alg-line:end-mlc} of \Cref{alg:mcr2-clustering}, `Data Augmentation' in \S \ref{sec:algo}). Below we give additional clarification on the role of augmentation therein. 
   
   \subsection{Augmentation for Initializing the Features} \label{sec:aug-for-init}
   
   Since the proposed \mours{} objective \eqref{eq:mcr2-clustering} is highly non-convex, the quality of the (local) solution an optimizer converges to in general depends on the initialization. However, before line \ref{alg-line:self-sup-init} of \Cref{alg:mcr2-clustering} is executed, the features $\mZ$ at initialization could be very far from union-of-orthogonal-subspaces (as pursued by Problem 1), since the neural network has an arbitrary architecture and initialization. To at least promote \textit{some} ideal structures in the features, line \ref{alg-line:self-sup-init} of \Cref{alg:mcr2-clustering} is conducted so that the features from an original sample and its augmented copies are close, while features from different samples spread out in the feature space. This is a common idea\footnote{More related are \cite[\S 3.2]{Yu2020-mx} and \cite[\S 3.6]{Li2022-vq} which are also based on \mcr{} as in this paper. However, the former does not learn a clustering membership, leading to inferior performance, and we discuss the difference with the latter in \S \ref{sec:related-work}, \ref{sec:mcr2-unsup-formulation} and \ref{sec:appendix-nmce}. } used in contrastive learning as we review in \S \ref{sec:related-work}.  Empirically, initializing the features using augmentation (line \ref{alg-line:self-sup-init} of \Cref{alg:mcr2-clustering}) is important for the clustering performance: as seen in \Cref{tab:ablation-study}, on CIFAR-10, if one uses random initialization to replace this step, then the final clustering accuracy is $20\%$, in sharp contrast to $86.3\%$.
   
   \subsection{Augmentation for Updating \bf{\mours{}} Objective \eqref{eq:mcr2-clustering}}
   
   In optimizing \mours{} loss \eqref{eq:mcr2-clustering} (lines \ref{alg-line:start-mlc}-\ref{alg-line:end-mlc} of \Cref{alg:mcr2-clustering}), augmentation empirically improves clustering performance. As one can see in \Cref{tab:ablation-study}, on CIFAR-10 using the sample self-supervised initialization of the features, \mours{} achieves only $80\%$ clustering accuracy without augmentation, in contrast to $86.3\%$ with augmentation. We attribute this difference to the fact that augmentation enriches the diversity of samples the algorithm sees.
   
   \section{Role and Ablation Study of Components of \Cref{alg:mcr2-clustering}}

   \subsection{Initialization of the Features (Line \ref{alg-line:self-sup-init})}
   Please kindly refer to \S \ref{sec:aug-for-init}. 
   
   \subsection{Updating the \bf{\mours{}} Objective \eqref{eq:mcr2-clustering} (Lines \ref{alg-line:start-mlc}-\ref{alg-line:end-mlc})}
   The main novelty of this paper lies in updating the \mours{} objective \eqref{eq:mcr2-clustering} that learns both the representation $\mZ$ and a doubly stochastic membership $\bGamma$. Note that in this step, clustering is pursued by modeling the membership $\bGamma$, as opposed to the self-supervised feature initialization step where no membership is explicitly pursued. This step is indeed important for clustering: as seen in \Cref{tab:ablation-study}, on CIFAR-10, the clustering accuracy on the self-supervised initialized features $\mZ$ is only $73.4\%$, in contrast to $86.3\%$ obtained after updating the \mours{} objective \eqref{eq:mcr2-clustering}. 
   
   \subsection{Spectral Clustering (Line \ref{alg-line: spectral-clustering})}
   Since the proposed \mours{} learns a doubly stochastic membership that signals pair-wise similarity between points, it is standard to run spectral clustering \cite{Von_Luxburg2007-xu} to compute a final set of clusters from the learned membership. This is done only once at the very end of \Cref{alg:mcr2-clustering}, and is rather efficient compared to the other parts of \Cref{alg:mcr2-clustering}: for instance, using an unaccelerated implementation from SciPy, it takes less than $30$ seconds to perform spectral clustering on a $10^4\times 10^4$ matrix. 
   
   \section{Complexity Analysis and Efficient Computation}
   A key observation is that the proposed \mours{} uses mini-batches (line 4 of Alg. 1). We summarize some important complexities in Table \ref{tab:my-table}.
\begin{table}[H]
    \centering
    \caption{Some key space / time complexities in computing $\log\det$ and Sinkhorn iterations, where $d$ is the dimension of the feature space and $n_b$ is the batch size. Real-data experiments in this paper use $n_b=1024$ and $d=128$.}
    \label{tab:my-table}
    \vspace{-0.2cm}
    \begin{tabular}{@{}llllll@{}}
    \toprule
      & $\log\det$ terms in (4) & Complexity   &  & Sinkhorn iteration & Complexity \\ \midrule
    1 & Size of matrix within $\log\det$ & $d\times d$ & 4 & Size of $\mC^\top \mC$ & $n_b\times n_b$ \\
    2 & Evaluate the matrix     & $O(n_b d^2)$ &  &                    &            \\
    3 & Evaluate the $\log\det$ & $O(d^3)$     &  &                    &            \\ \bottomrule
    \end{tabular}%
\end{table}
\noindent First, computations 1-3 have the same complexity as in previous works \cite{yu2020learning, Li2022-vq}. The only extra overhead comes from the sum in the $R_c$ term in (4) being of size $n_b$ rather than $k$, where $k$ is \# of clusters. However, empirically the extra overhead is negligible using buit-in pararallization from \texttt{torch.einsum}. Second, scalability of Sinkhorn iterations on a $n_b\times n_b$ matrix is also not a concern. Since\footnote{As is also pointed out in \S \ref{sec:algo}.}, in all real-data experiments, while $n\sim 10^5$ is typically very large, $n_b=1024$ is much smaller than $n$. Thus, the time and space complexities remain manageable despite the datasets being large-scale.
Indeed, as seen in \Cref{tab:exp-cifar100-timgnet200}, \mours{} achieves higher accuracy in less time than state-of-the-art alternatives on CIFAR100-100. Finally, we note that
efficient computation of $\log\det$ is a subject of research (e.g., \cite{Baek2022-bj}), and back propagation of Sinkhorn iterations can be made efficient using implicit differentiation \cite{Eisenberger2022-oo}; these are beyond the scope of this paper.

   \begin{table*}[]
   \centering
   \caption{Clustering accuracy and normalized mutual information of \mours{} and \texttt{NMCE} \cite{Li2022-vq} on CIFAR-10 over $10$ random seeds, using the same self-initialized features. Note that \mours{} is more accurate and stable than \texttt{NMCE}, which is attributed to the fact that the doubly stochastic membership of \mours{} can be deterministically initialized using self-supervised features (\S \ref{sec:algo}).}
   \label{tab:seed-mlc-nmce}
   \resizebox{0.85\textwidth}{!}{%
   \begin{tabular}{@{}lllcccccccccclll@{}}
   \toprule
   \multirow{2}{*}{Method} &  & \multirow{2}{*}{Metric} & \multicolumn{10}{c}{Seed} &  & \multirow{2}{*}{Mean} & \multirow{2}{*}{Std.} \\ \cmidrule(lr){4-13}
    &  &  & 0 & 1 & 2 & 3 & 4 & 5 & 6 & 7 & 8 & 9 &  &  &  \\ \midrule
   \multirow{2}{*}{\mours{} \eqref{eq:mcr2-clustering}} &  & ACC & 84.5 & 84.8 & 84.8 & 84.6 & 84.4 & 84.4 & 84.0 & 84.3 & 84.4 & 84.6 &  & \textbf{84.5} & \textbf{0.24} \\
    &  & NMI & 76.6 & 77.1 & 76.8 & 76.8 & 76.5 & 76.4 & 76.1 & 76.4 & 76.4 & 76.5 &  & \textbf{76.6} & \textbf{0.28} \\ \midrule
   \multirow{2}{*}{\texttt{NMCE}} &  & ACC & \multicolumn{1}{l}{83.7} & 82.1 & 81.6 & 73.7 & 80.4 & 77.9 & 81.7 & 81.4 & 72.7 & 80.9 &  & 79.6 & 3.69 \\
    &  & NMI & \multicolumn{1}{l}{74.4} & 71.2 & 70.4 & 65.2 & 70.0 & 68.1 & 72.7 & 70.8 & 69.2 & 69.8 &  & 70.2 & 2.49 \\ \bottomrule
   \end{tabular}%
   }
   \end{table*}
   
   \section{Additional Comparison of \bf{\mours{}} and \bf{\texttt{NMCE}}}\label{sec:appendix-nmce}
   As detailed in \S \ref{sec:mcr2-unsup-formulation}, one of the advantages of the proposed \mours{} \eqref{eq:mcr2-clustering} over \texttt{NMCE} \cite{Li2022-vq} is that \mours{} has a more stable performance with respect to random seeds, since \mours{} is able to initialize the membership deterministically using structures from the self-supervised initialized features. Below we conduct extra experiments to provide empirical evidence. We first fix a self-supervised initialization of features that is in turn used for both \texttt{NMCE} and \mours{}. Then, based on this very same initialization of features, we update \texttt{NMCE} and \mours{} objective respectively with $10$ different seeds: recall that \texttt{NMCE} initializes the membership randomly whereas \mours{} initializes the membership deterministically using the initialized features. To make a valid comparison, for both methods we further use the same optimization strategy and hyper-parameters that are optimally\footnote{For \texttt{NMCE}, we use the implementation as well as the parameters provided in \scriptsize{\url{https://github.com/zengyi-li/NMCE-release}}.} tuned for \texttt{NMCE} (which are not optimal for \mours{}): precision $\epsilon^2=0.2$, \# epochs $100$, LARS optimizer for $\mZ$ with an initial learning rate $0.3$ decayed to $0$ in a cosine annealing manner. \Cref{tab:seed-mlc-nmce} reports clustering accuracy and normalized mutual information of \mours{} and \texttt{NMCE} over $10$ random seeds. As expected, \mours{} has a more stable clustering performance by having a standard deviation of clustering accuracy and normalized mutual information less than $0.28$, in contrast to more than $2.49$ achieved by \texttt{NMCE}. Further, \mours{} achieves higher mean clustering performance than \texttt{NMCE}, as also observed in \Cref{tab:exp-cifar10-20}. Last but not least, we note that the numbers in \Cref{tab:seed-mlc-nmce} are not comparable to those in \Cref{tab:exp-cifar10-20}, since for \mours{} the hyper-parameters and optimizers are different, and for \texttt{NMCE} an additional step that fine tunes the backbone is used in \Cref{tab:exp-cifar10-20}.

   \section{Addtional Visualization on Learned Representation and Clusters}
\begin{figure*}
    \centering
    \begin{subfigure}{0.33\textwidth}
      \includegraphics[width=\linewidth,trim={1cm 0 1cm 0.8cm},clip]{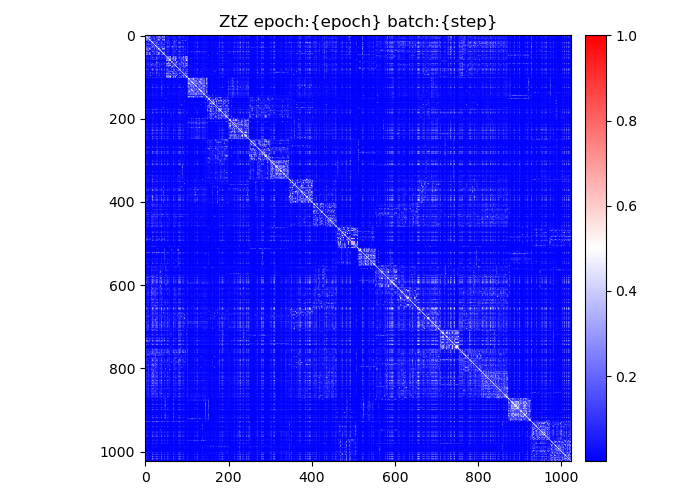}
      \caption{CIFAR-20}
      \label{fig:CIFAR-20-heat}
    \end{subfigure}
    \hfill
    \begin{subfigure}{0.33\textwidth}
      \includegraphics[width=\linewidth,trim={1cm 0 1cm 0.8cm},clip]{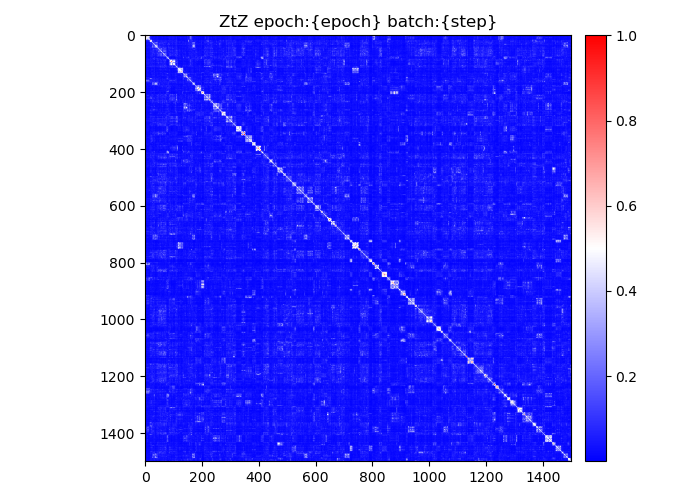}
      \caption{CIFAR-100}
      \label{fig:CIFAR-100-heap}
    \end{subfigure}
    \hfill
    \begin{subfigure}{0.33\textwidth}
      \includegraphics[width=\linewidth,trim={1cm 0 1cm 0.8cm},clip]{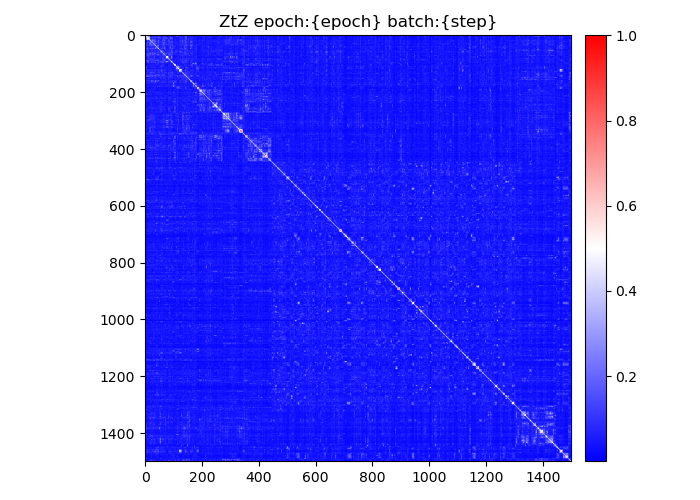}
      \caption{TinyImageNet-200}
      \label{fig:Tiny_ImageNet-heat}
    \end{subfigure}
    \caption{Cosine similarity $|\mZ^\top \mZ|$ of the features $\mZ$ learned by \mours{} on more complicated datasets: CIFAR-20, CIFAR-100, TinyImageNet-200.}
    \label{fig:more_heatmap}
  \end{figure*}

\begin{figure*}
     
     \begin{subfigure}{0.32\textwidth}
     \addtocounter{subfigure}{12}
       \includegraphics[width=\linewidth,trim={1cm 2cm 1cm 2cm},clip]{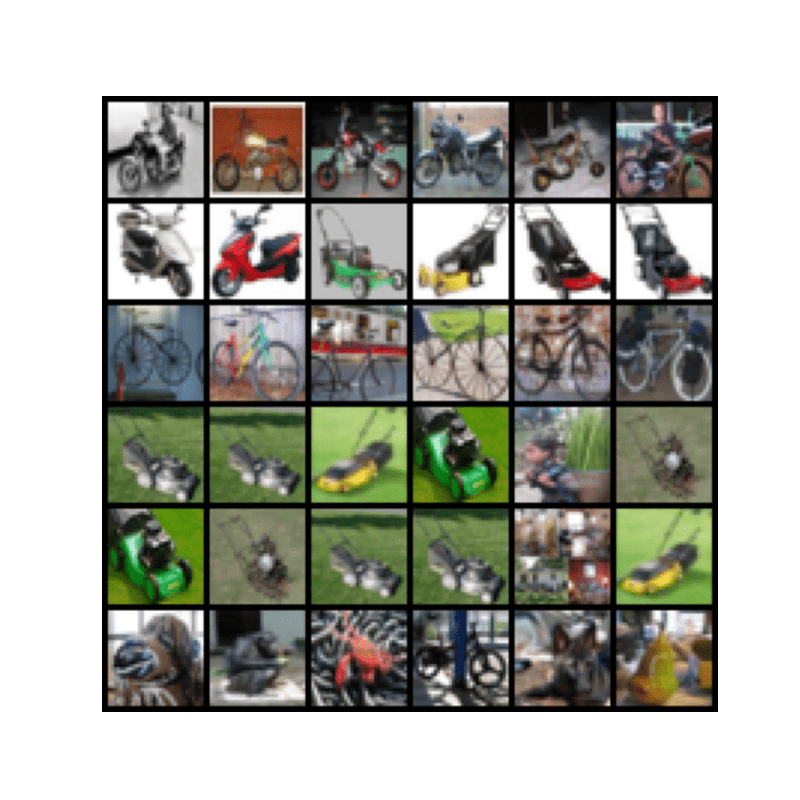}
       \caption{Learned cluster 13}
     \end{subfigure}
     \hfill
     \begin{subfigure}{0.32\textwidth}
       \includegraphics[width=\linewidth,trim={1cm 2cm 1cm 2cm},clip]{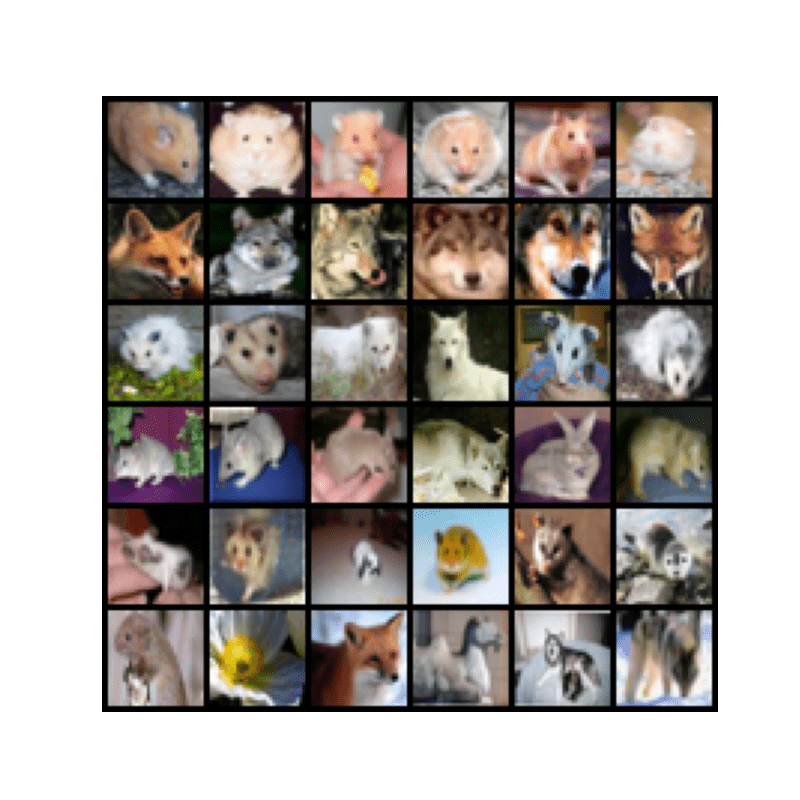}
       \caption{Learned cluster 14}
     \end{subfigure}
     \hfill
     \begin{subfigure}{0.32\textwidth}
       \includegraphics[width=\linewidth,trim={1cm 2cm 1cm 2cm},clip]{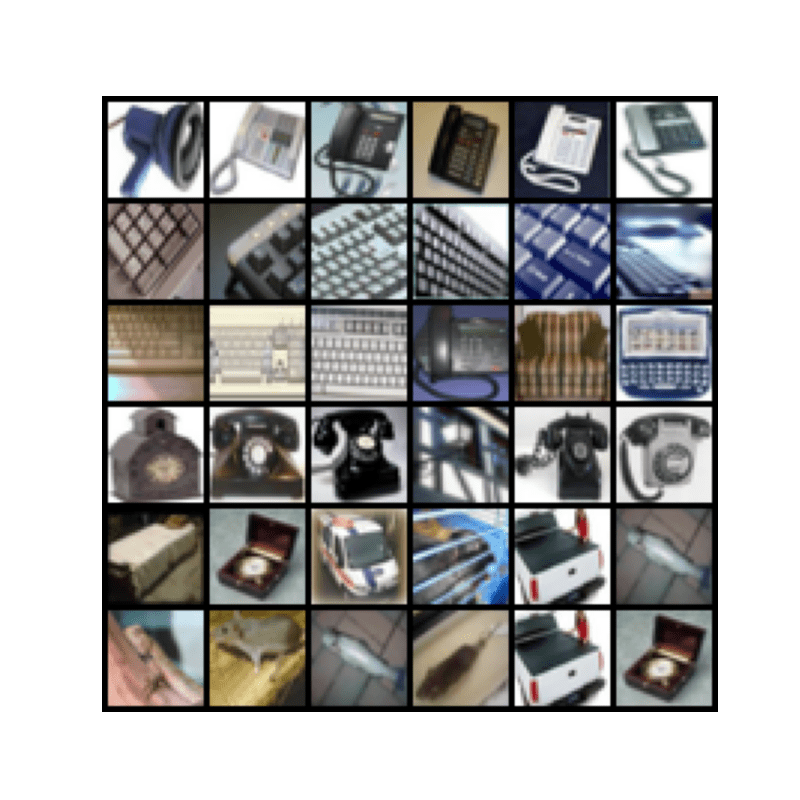}
       \caption{Learned cluster 15}
     \end{subfigure}
     \hfill
     \begin{subfigure}{0.32\textwidth}
       \includegraphics[width=\linewidth,trim={1cm 2cm 1cm 2cm},clip]{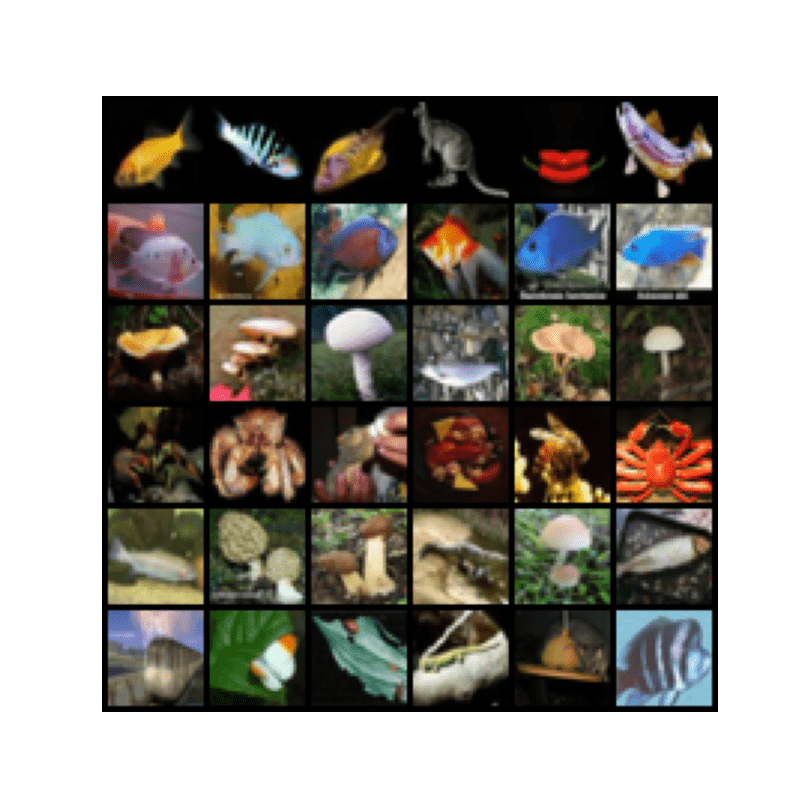}
       \caption{Learned cluster 16}
     \end{subfigure}
     \hfill
     \begin{subfigure}{0.32\textwidth}
       \includegraphics[width=\linewidth,trim={1cm 2cm 1cm 2cm},clip]{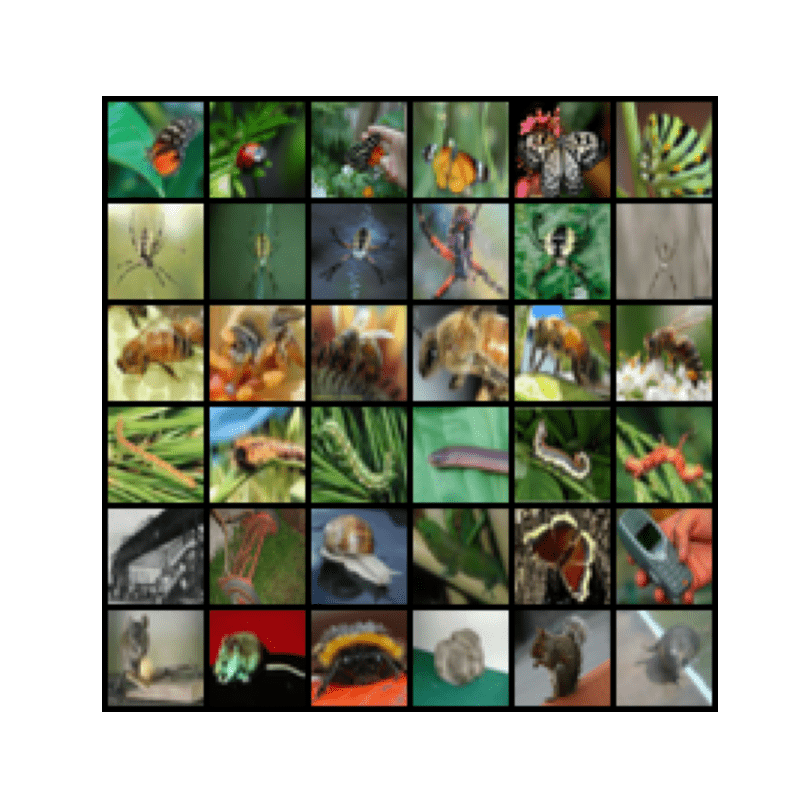}
       \caption{Learned cluster 17}
     \end{subfigure}
     \hfill
     \begin{subfigure}{0.32\textwidth}
       \includegraphics[width=\linewidth,trim={1cm 2cm 1cm 2cm},clip]{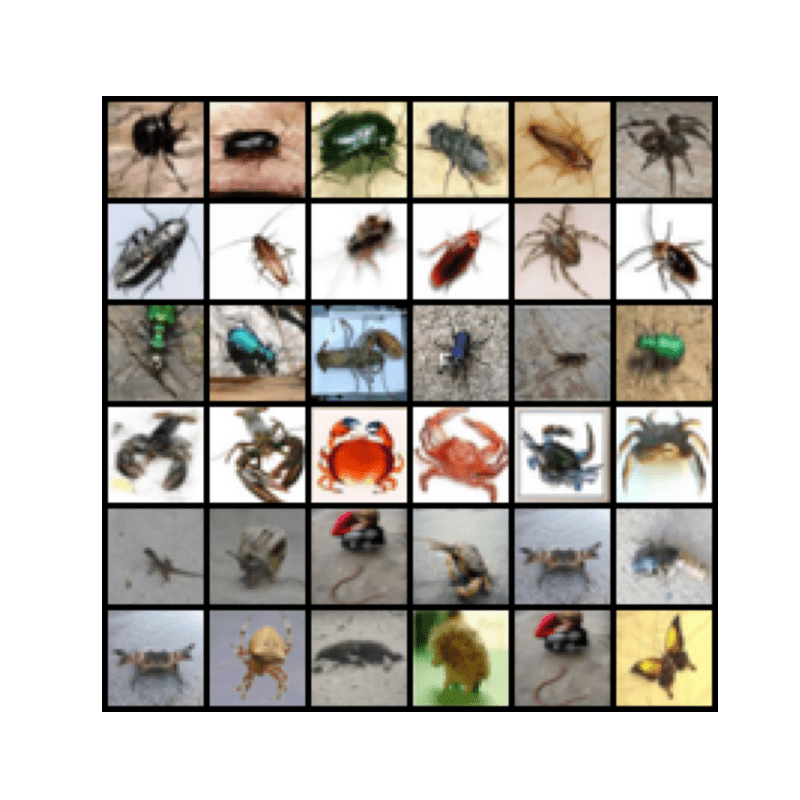}
       \caption{Learned cluster 18}
     \end{subfigure}
     \hfill
     \begin{subfigure}{0.32\textwidth}
       \includegraphics[width=\linewidth,trim={1cm 2cm 1cm 2cm},clip]{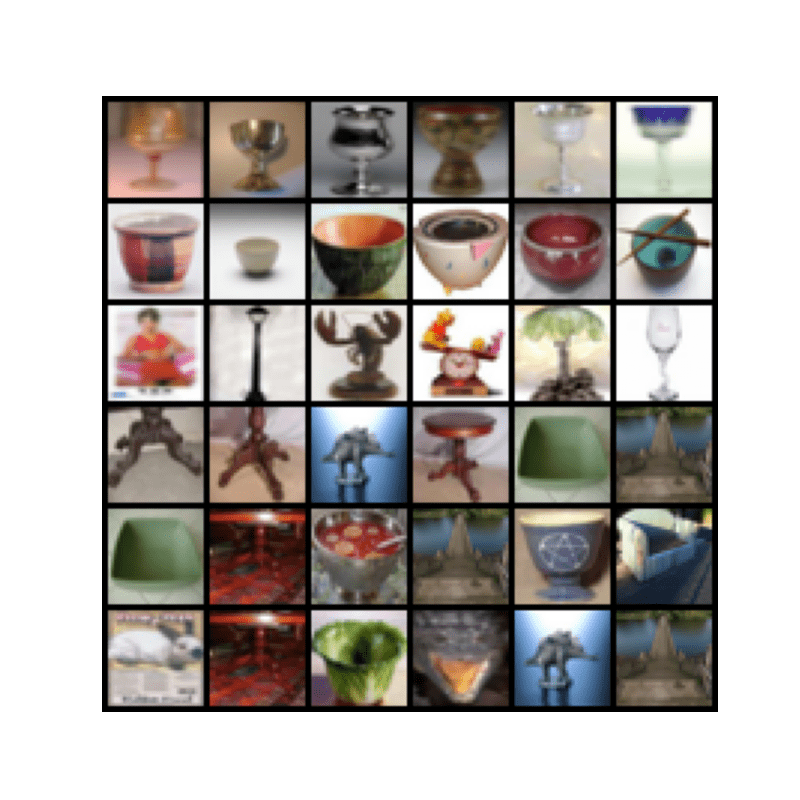}
       \caption{Learned cluster 19}
     \end{subfigure}
     \hfill
     \begin{subfigure}{0.32\textwidth}
       \includegraphics[width=\linewidth,trim={1cm 2cm 1cm 2cm},clip]{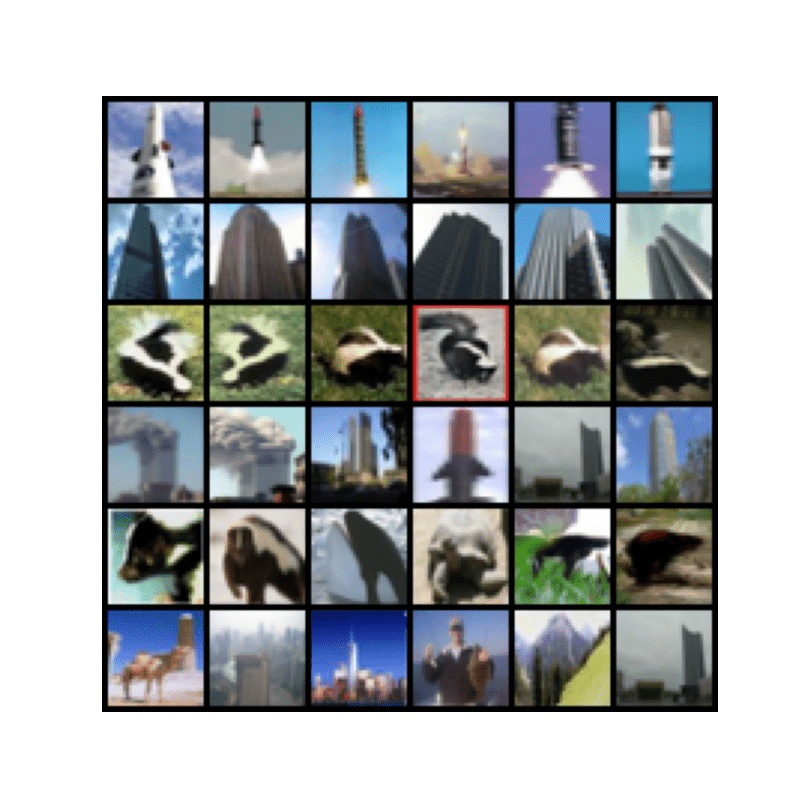}
       \caption{Learned cluster 20}
     \end{subfigure}
     \caption{Images along the principal components (defined in \S \ref{sec:semantic}) of features from each cluster on CIFAR-20, where features and clusters are learned by \eqref{eq:mcr2-clustering}.}
     \label{fig:cifar100-20-pc}
   \end{figure*}

   \Cref{fig:more_heatmap} presents the cosine similarity (as defined in the preamble of \S \ref{sec:exp}) of the representation learned by \mours{} on CIFAR-20, CIFAR-100 and TinyImageNet-200 (for the counterpart on CIFAR-10 see \S \ref{sec:exp-sc}). As seen, the cosine similarity maps form approximately block diagonal structures, showing that the features from different clusters are roughly orthogonal to each other. This is desired by the between-cluster discrimination (\S \ref{sec:intro}). Lastly, we provide additional visualization of principal images on CIFAR-20 (see \S \ref{sec:semantic} for definition) in \Cref{fig:cifar100-20-pc}. 

   \begin{figure*}
     \centering
     \begin{subfigure}{0.32\textwidth}
       \includegraphics[width=\linewidth,trim={1cm 2cm 1cm 2cm},clip]{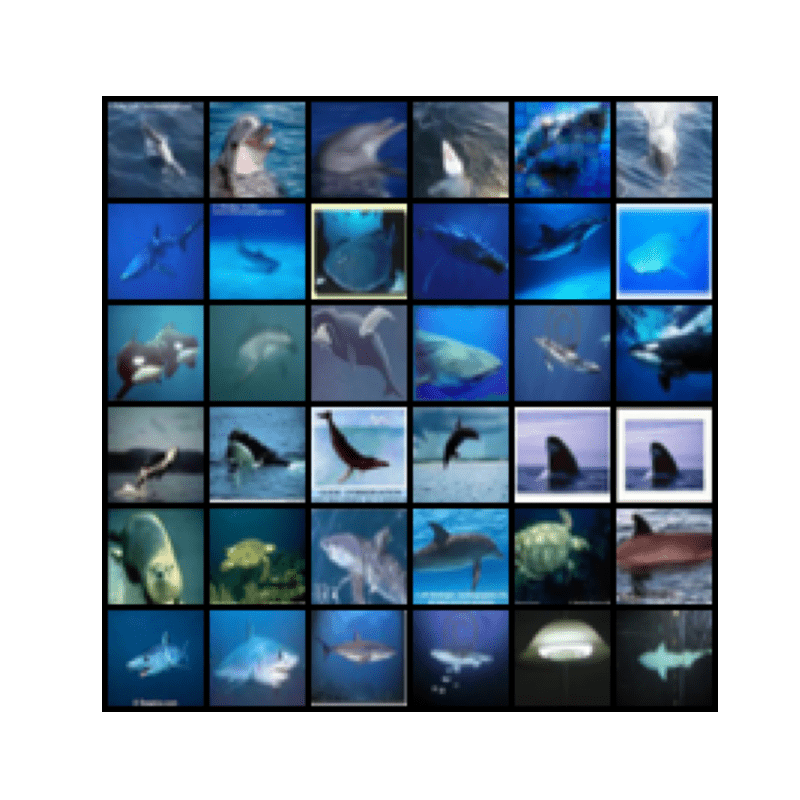}
       \caption{Learned cluster 1}
     \end{subfigure}
     \hfill
     \begin{subfigure}{0.32\textwidth}
       \includegraphics[width=\linewidth,trim={1cm 2cm 1cm 2cm},clip]{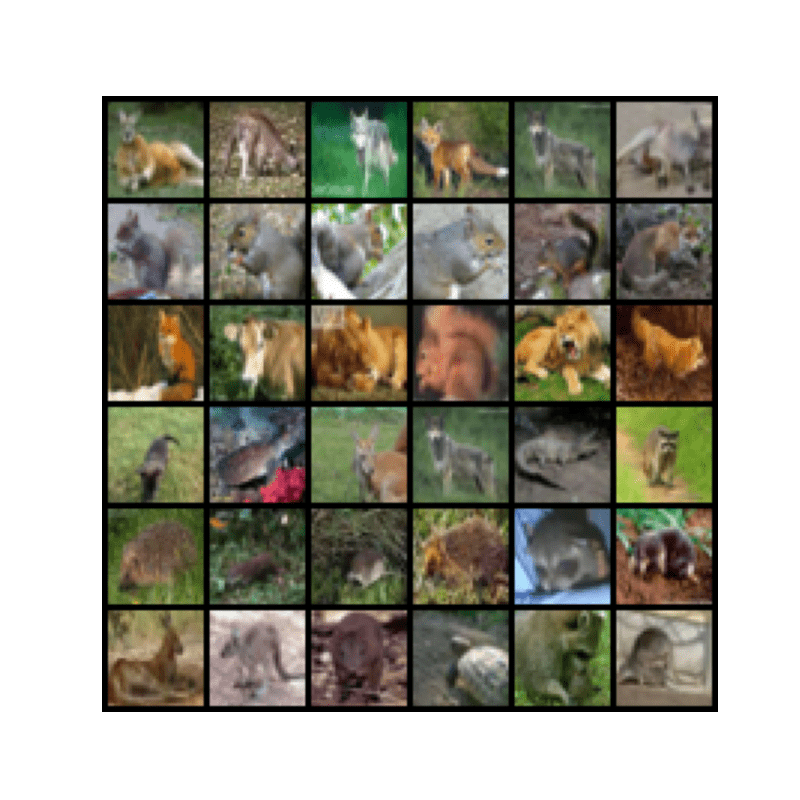}
       \caption{Learned cluster 2}
     \end{subfigure}
     \hfill
     \begin{subfigure}{0.32\textwidth}
       \includegraphics[width=\linewidth,trim={1cm 2cm 1cm 2cm},clip]{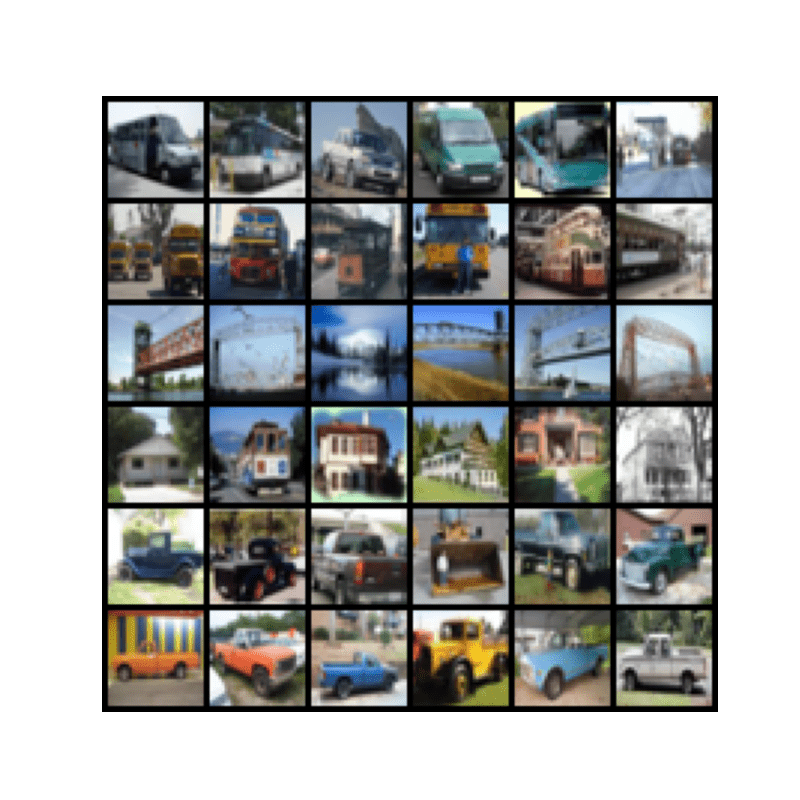}
       \caption{Learned cluster 3}
     \end{subfigure} 
     \hfill
     \begin{subfigure}{0.32\textwidth}
       \includegraphics[width=\linewidth,trim={1cm 2cm 1cm 2cm},clip]{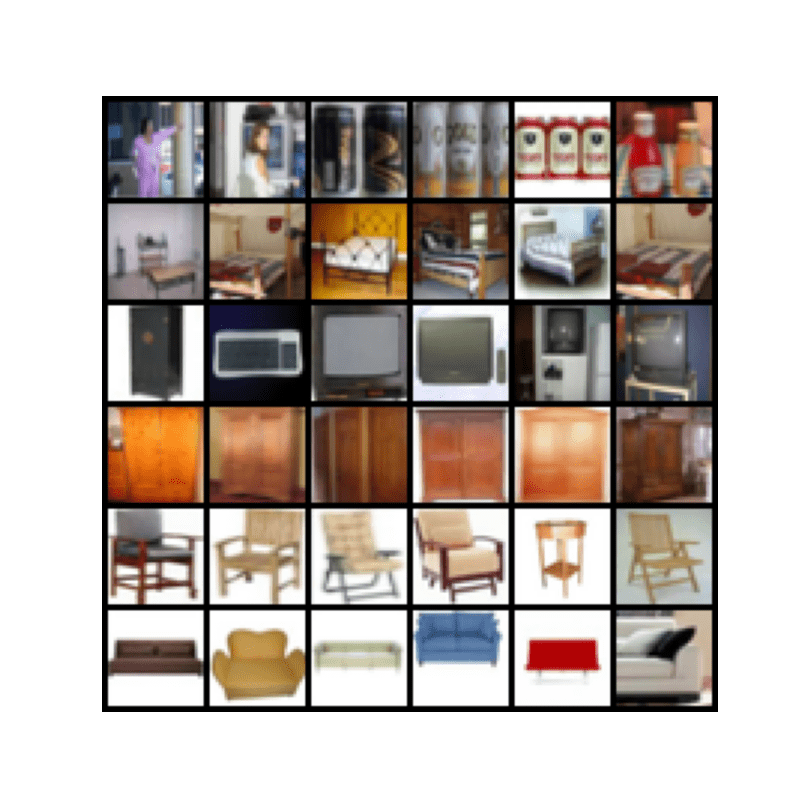}
       \caption{Learned cluster 4}
     \end{subfigure}
     \hfill
     \begin{subfigure}{0.32\textwidth}
       \includegraphics[width=\linewidth,trim={1cm 2cm 1cm 2cm},clip]{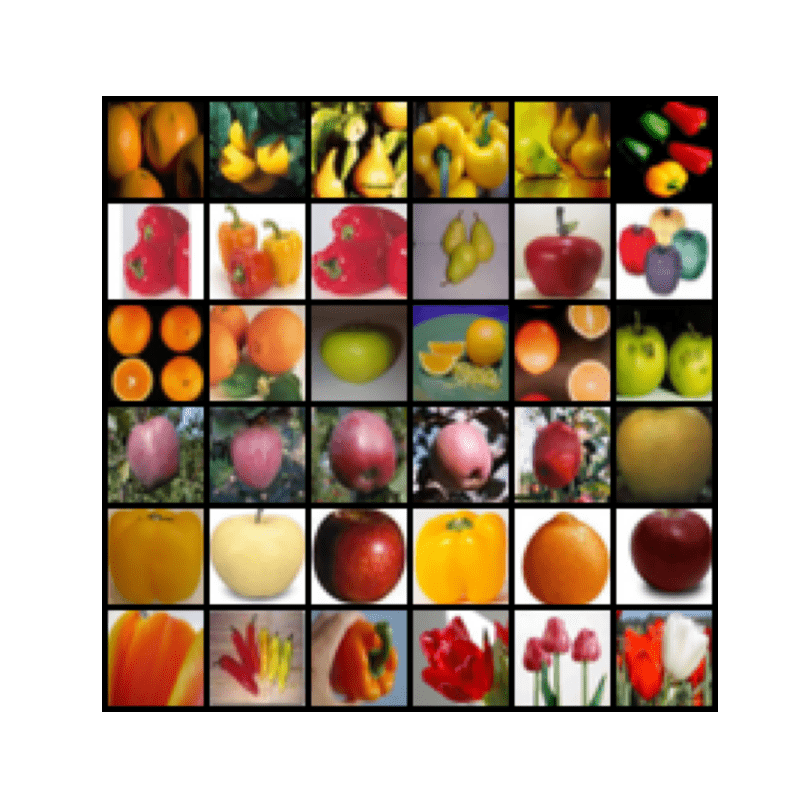}
       \caption{Learned cluster 5}
     \end{subfigure}
     \hfill
     \begin{subfigure}{0.32\textwidth}
       \includegraphics[width=\linewidth,trim={1cm 2cm 1cm 2cm},clip]{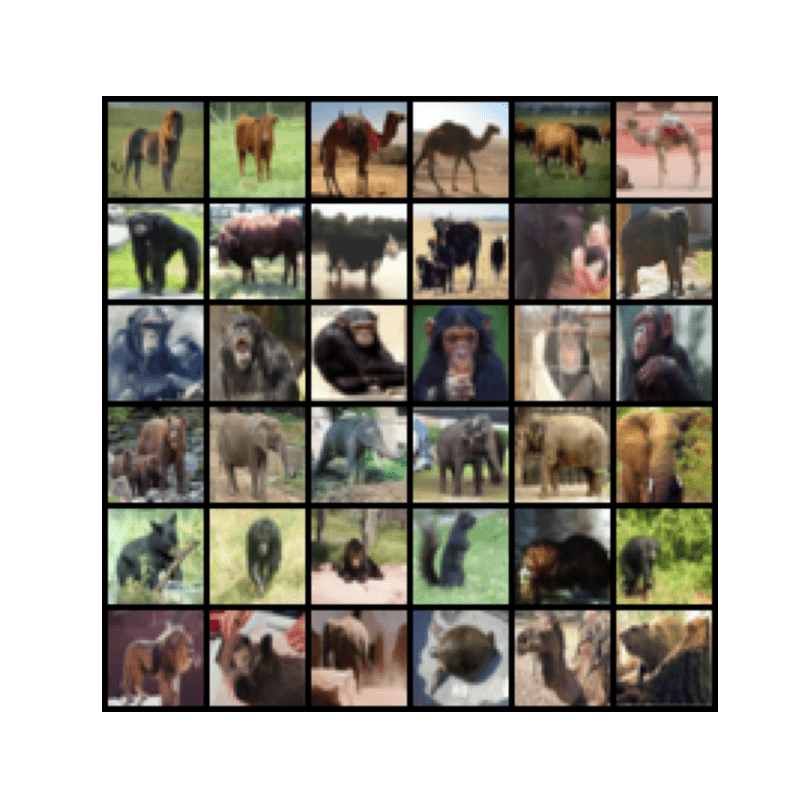}
       \caption{Learned cluster 6}
     \end{subfigure}
     \hfill
     \begin{subfigure}{0.32\textwidth}
       \includegraphics[width=\linewidth,trim={1cm 2cm 1cm 2cm},clip]{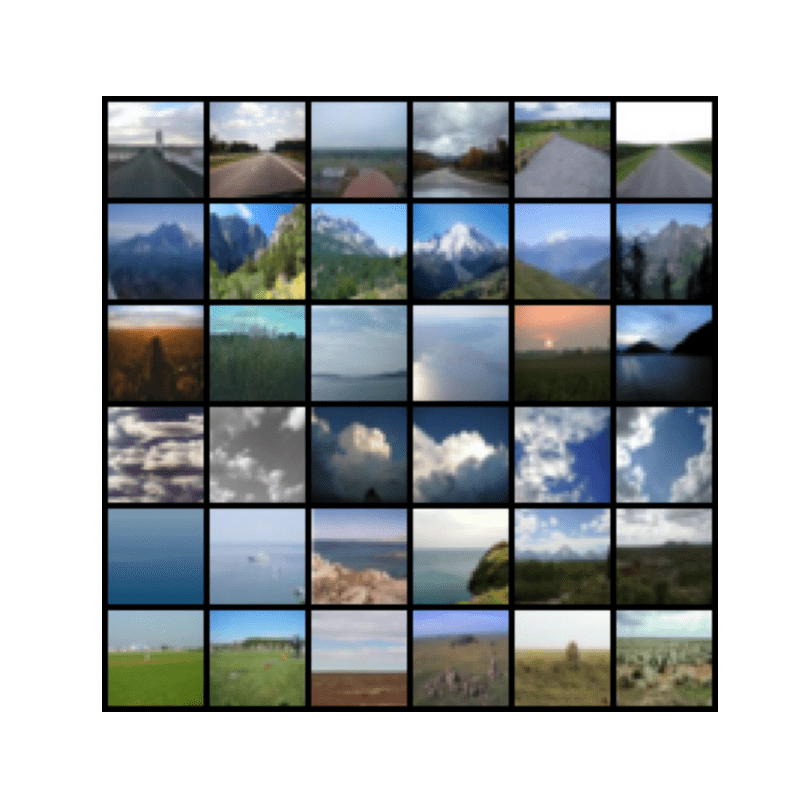}
       \caption{Learned cluster 7}
     \end{subfigure}
     \hfill
     \begin{subfigure}{0.32\textwidth}
       \includegraphics[width=\linewidth,trim={1cm 2cm 1cm 2cm},clip]{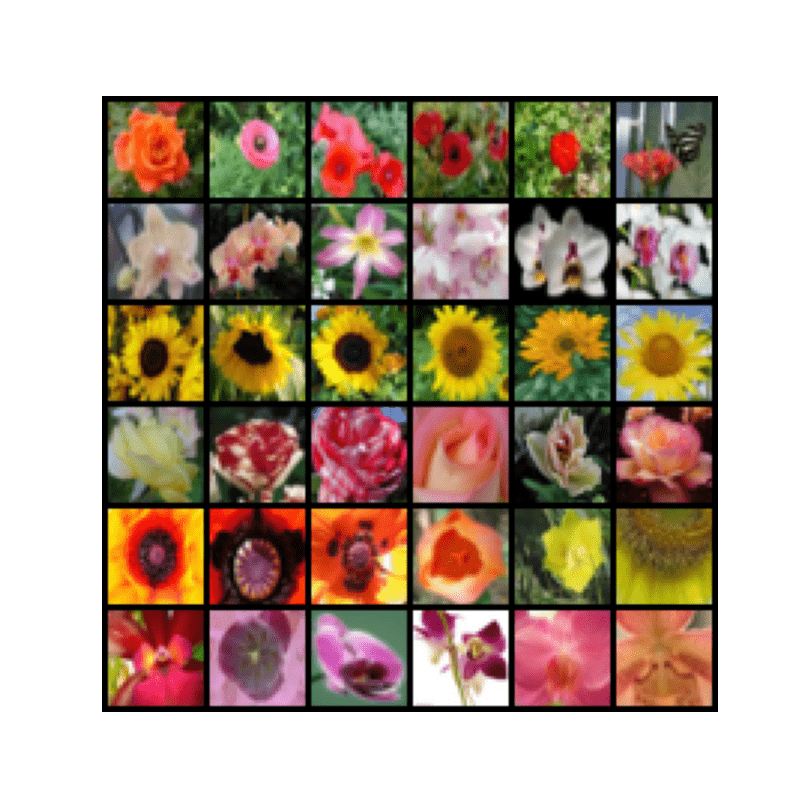}
       \caption{Learned cluster 8}
     \end{subfigure}
     \hfill
     \begin{subfigure}{0.32\textwidth}
       \includegraphics[width=\linewidth,trim={1cm 2cm 1cm 2cm},clip]{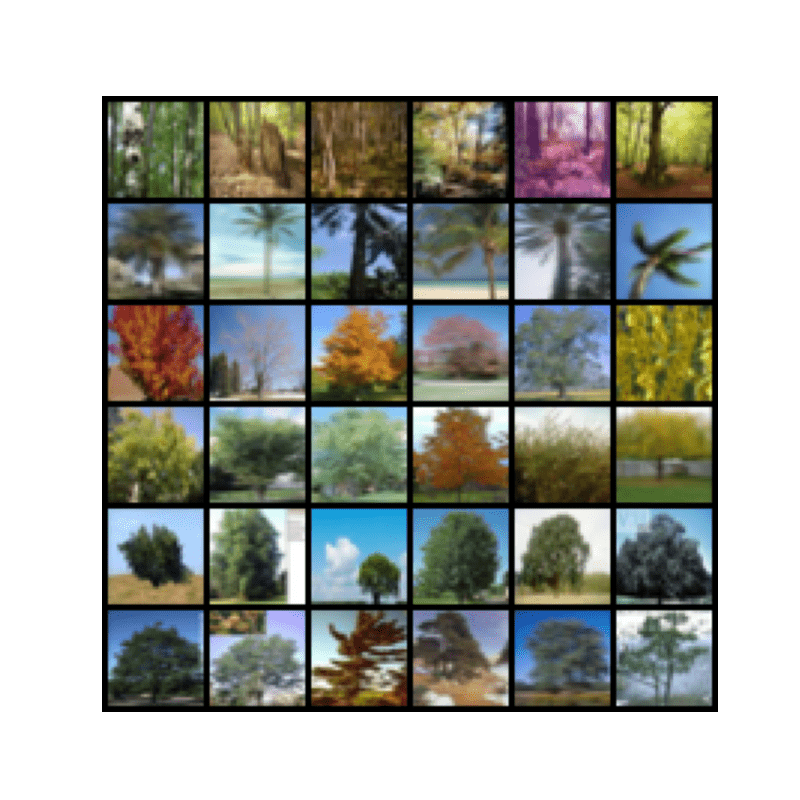}
       \caption{Learned cluster 9}
     \end{subfigure}
     \hfill
     \begin{subfigure}{0.32\textwidth}
       \includegraphics[width=\linewidth,trim={1cm 2cm 1cm 2cm},clip]{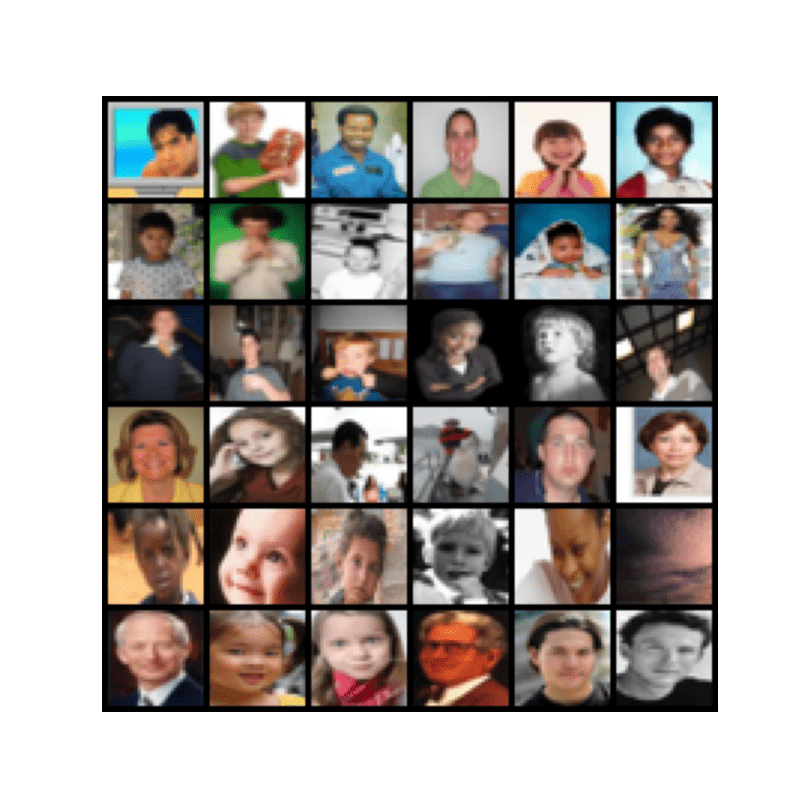}
       \caption{Learned cluster 10}
     \end{subfigure}
     \hfill
     \begin{subfigure}{0.32\textwidth}
       \includegraphics[width=\linewidth,trim={1cm 2cm 1cm 2cm},clip]{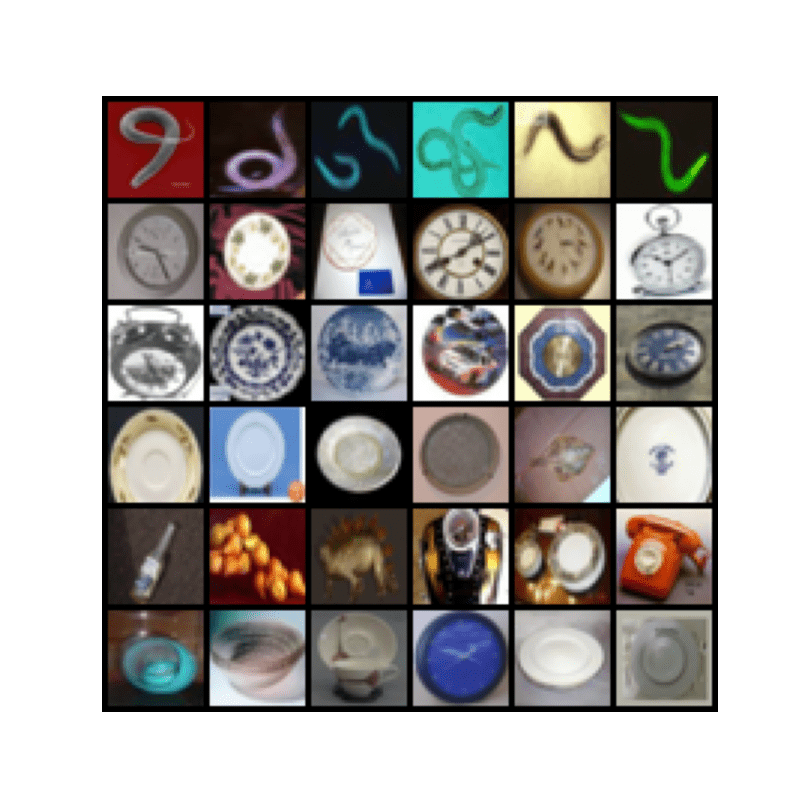}
       \caption{Learned cluster 11}
     \end{subfigure}
     \hfill
     \begin{subfigure}{0.32\textwidth}
       \includegraphics[width=\linewidth,trim={1cm 2cm 1cm 2cm},clip]{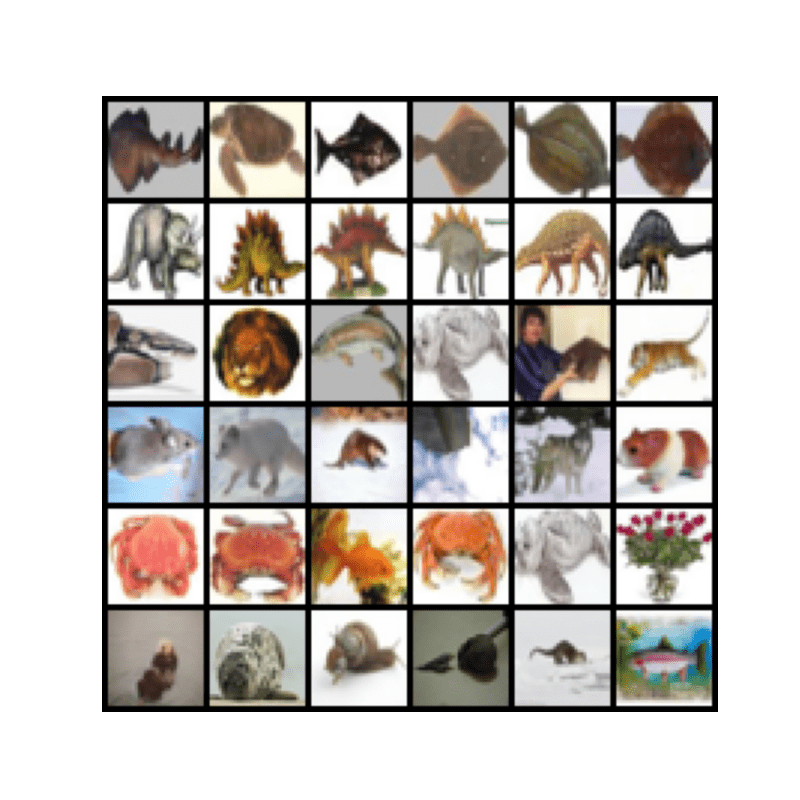}
       \caption{Learned cluster 12}
     \end{subfigure}
   \end{figure*}  

\end{document}